%% file: main.tex
\documentclass[10pt,twocolumn,letterpaper]{article}
\usepackage[pagenumbers]{cvpr}              %

\input{tex/preamble}

\input{tex/hyperref}

\input{tex/notation}

\def\confName{CVPR}
\def\confYear{2027}

\begin{document}

\input{tex/title.tex}

\input{tex/abstract.tex}
\input{tex/introduction.tex}
\input{tex/method.tex}
\input{tex/experiments.tex}
\input{tex/defenses.tex}
\input{tex/conclusion.tex}
\clearpage
\input{tex/acknowledgments.tex}
\input{tex/bibliography}
\clearpage
\input{tex/suppl.tex}

\end{document}

%% file: tex/preamble.tex
\usepackage{array}
\usepackage{epstopdf}
\usepackage{multirow}
\usepackage{amsthm}
\usepackage{dsfont}
\usepackage{algorithm}
\usepackage{algorithmic}
\usepackage{tabularx}
\usepackage{arydshln} %

\usepackage[hang,flushmargin]{footmisc}
\usepackage{microtype} %

\newcolumntype{Y}{>{\centering\arraybackslash}X}
\newcolumntype{R}{>{\raggedleft\arraybackslash}X}
\newcolumntype{L}{>{\raggedright\arraybackslash}X}
\newcolumntype{M}{>{\centering\arraybackslash$}X<{$}}  %

\renewcommand{\eqref}[1]{\hyperref[#1]{(}\ref{#1}\hyperref[#1]{)}}

\newcommand{\suppref}[1]{Supp.~\ref{#1}}

\newcommand{\suppfrozen}[1]{\ifnum\value{section}=#1\relax\else
  \PackageError{suppl}{Supplement section \thesection\space is frozen as S#1:
  the submitted main paper cites it by that number}{Do not add, remove or
  reorder supplement sections before the last \string\suppfrozen.}\fi}

\makeatletter
\@ifundefined{linenumbers}{}{%
  \RequirePackage{zref-savepos}
  \newcounter{capbox}
  \newcommand\capline@prefix{%
    \def\@tmpa{figure}\ifx\@captype\@tmpa F\thefigure\else T\thetable\fi.}
  \DeclareCaptionFormat{linenumbered}{%
    \stepcounter{capbox}%
    \zsavepos{capbox\thecapbox}%
    \begingroup
    \ifnum\zposx{capbox\thecapbox}>\dimexpr0.5\paperwidth\relax
      \rightlinenumbers*\else\leftlinenumbers*\fi
    \renewcommand\thelinenumber{\capline@prefix\arabic{linenumber}}%
    \internallinenumbers*#1#2#3\par
    \endgroup}
  \captionsetup{format=linenumbered}
  \captionsetup[sub]{format=plain}
}
\makeatother

\newcommand\customparagraph[1]{\vspace{0em}\noindent\textbf{#1}}

\definecolor{figmapblue}{HTML}{3987E5}   %
\definecolor{figplacement}{HTML}{D62828} %
\definecolor{figorange}{HTML}{EB6834}    %
\definecolor{figink}{HTML}{0B0B0B}
\newcommand{\legendglyph}[2]{{\setlength{\unitlength}{1pt}\color{#1}\raisebox{-0.6pt}{\begin{picture}(5.5,5.5)#2\end{picture}}}}
\newcommand{\swatchmap}{\legendglyph{figmapblue}{\put(0,0){\rule{5.5pt}{5.5pt}}}}
\newcommand{\swatchring}{\legendglyph{figorange}{\linethickness{0.7pt}\put(2.75,2.75){\circle{5.5}}}}
\newcommand{\swatchsolid}{\legendglyph{figink}{\linethickness{0.6pt}\put(0,0){\framebox(5.5,5.5){}}}}
\newcommand{\swatchdashed}{\legendglyph{figplacement}{\linethickness{0.6pt}\put(0,0){\dashbox{1.1}(5.5,5.5){}}}}

\definecolor{cblue}{rgb}{0, 0.44, 0.75}
\definecolor{cred}{rgb}{0.75, 0, 0}
\definecolor{cgreen}{rgb}{0, 0.55, 0.3}

%% file: tex/hyperref.tex
\definecolor{cvprblue}{rgb}{0.21,0.49,0.74}
\usepackage[pagebackref,breaklinks,colorlinks,allcolors=cvprblue]{hyperref}

\usepackage{etoc}
\usepackage{tocloft}
\AtBeginDocument{\etocdepthtag.toc{main}}
\etocsettagdepth{main}{none}
\etocsettagdepth{supp}{subsection}

\makeatletter\renewcommand{\@tocrmarg}{2.55em plus 1fil}\makeatother

\makeatletter
\def\cvprsect#1{\cvprsection[#1]{\texorpdfstring{\hskip -1em.~}{}#1}}
\def\cvprsubsect#1{\cvprsubsection[#1]{\texorpdfstring{\hskip -1em.~}{}#1}}
\def\cvprsubsubsect#1{\cvprsubsubsection[#1]{\texorpdfstring{\hskip -1em.~}{}#1}}
\makeatother

%% file: tex/notation.tex
\newcommand{\pos}{\boldsymbol{x}}          %
\newcommand{\Pos}{\boldsymbol{X}}          %
\newcommand{\posrec}{\hat{\boldsymbol{x}}} %
\newcommand{\Posrec}{\hat{\boldsymbol{X}}} %

\newcommand{\descr}{\boldsymbol{d}}        %
\newcommand{\Descr}{\mathcal{D}}           %
\newcommand{\dimd}{D}                       %

\newcommand{\off}{\boldsymbol{o}}          %
\newcommand{\numkp}{N}                      %

\newcommand{\Center}{\boldsymbol{C}}

\newcommand{\Real}{\mathds{R}}

\newcommand{\solver}{f_{\boldsymbol{\theta}}} %

%% file: tex/title.tex
\title{DescOver: Descriptor Overlap Leaks the Image}

\author{Simona Kocour$^{1,2}$ \,\, Torsten Sattler$^{1}$ \,\, Zuzana Kukelova$^{3}$ \,\,Marc Pollefeys$^{4,5}$ \,\, Pablo Speciale$^{4}$\vspace*{0.5ex}\\ 
\small $^1$ Czech Institute of Informatics, Robotics and Cybernetics, Czech Technical University in Prague \\
\small $^2$ Faculty of Electrical Engineering, Czech Technical University in Prague \\
\small $^3$ Visual Recognition Group, Faculty of Electrical Engineering, Czech Technical University in Prague \\
\small $^4$ ETH Zurich \quad
$^5$ Microsoft Spatial AI Lab, Zurich 
}

\makeatletter
\patchcmd{\@maketitle}{\vskip .375in}{\vskip .04in}{}{}
\patchcmd{\@maketitle}{\vspace*{24pt}}{\vspace*{15pt}}{}{\PatchFailed}
\makeatother
\twocolumn[{
\renewcommand\twocolumn[1][]{#1}
\maketitle
\vspace{-20pt}
\begin{center}
  \captionsetup{type=figure}
  \includegraphics[width=1.0\textwidth,trim=5.5 0 5.5 0,clip]{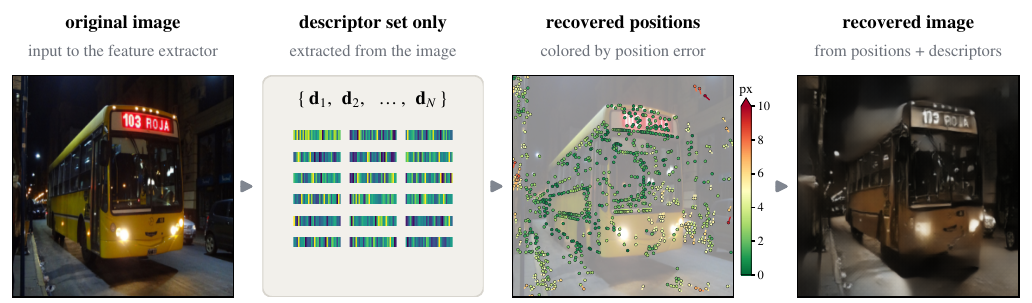}
  \captionof{figure}{\textbf{Can an image be reconstructed from its descriptors alone?} Recognizable image content can be reconstructed from local features, \ie, descriptors and their 2D keypoint positions. We show that 2D keypoints are not necessarily needed for the recovery process. From descriptors alone, our trained transformer predicts one 2D position per descriptor. An inversion model then reconstructs recognizable content from the descriptors at these predicted positions.
  }
  \label{fig:teaser}
\end{center}
}]

%% file: tex/abstract.tex
\vspace*{-2ex}

\begin{abstract}
Recognizable image content can be reconstructed from keypoint positions and descriptors of local features.
This creates potential privacy risks when transmitting such data, \eg, in cloud-based visual localization.
Obfuscation schemes aim to prevent these inversion attacks while retaining utility for tasks such as camera pose estimation.
In this paper, we study the limiting case of geometric obfuscation, where all keypoint positions are discarded.
Most modern learned descriptors are typically upright, \ie, computed without rotating image regions to per-keypoint orientations, and for these we show that \textbf{keypoint positions can be recovered from the descriptors alone}, enabling image inversion.
Our key observation is that neighboring descriptors describe overlapping image regions in a common image frame, so their shared content reveals where their keypoints are relative to each other, not only which keypoints are neighbors.
A set transformer trained on public images exploits this shared content to jointly predict all keypoint positions.
We recover keypoint positions for five learned feature extractors, and the trained solvers also work on image domains unseen during training.
Recovery also persists in several tested settings after training on subsampled or quantized descriptors.
\textbf{These results expose a limit of geometric obfuscation}: for the descriptors studied, recognizable image content can be reconstructed even with all keypoint positions discarded.
We will make our code and trained models publicly available.

\end{abstract}

%% file: tex/introduction.tex
\vspace{-4pt}

\section{Introduction}
\label{sec:intro}

Many applications require users to send image data to external services, \eg, cloud-based 3D reconstruction~\cite{agarwal2009rome,geppert2020privacy} or client-server-based visual localization~\cite{lynen2020largescale,speciale2019queries}. 
Sending images leads to potential privacy risks. 
Thus, it is common to extract features from the images and send the features instead of the original image.
For example, in the context of visual localization, a client might send local image features to a server, where they are used to estimate the camera pose \wrt a given 3D model of the scene~\cite{speciale2019queries}. 
Each local feature consists of a 2D keypoint position and a descriptor of the surrounding image content. 
However, inversion attacks can reconstruct recognizable image content from the positions and descriptors of sparse features~\cite{pittaluga2019revealing,pietrantoni2026vulnerability}.

Inversion attacks can potentially be prevented by obfuscating the keypoint positions~\cite{speciale2019queries,geppert2020privacy,geppert2021privacy,geppert2022partial,Pan_2023_ICCV,lee2023pairedpoint,moon2024rayclouds,kim2026dcl} or descriptors~\cite{dusmanu2021privacy,ng2022ninjadesc,pittaluga2023ldpfeat,pietrantoni2023segloc,panek2026structureless}. 
Geometric obfuscation  %
introduces ambiguities in the positions, \eg, %
by replacing each 2D  position with a random line through it~\cite{speciale2019queries}. 
Descriptor obfuscation introduces ambiguities into the descriptors such that they do not carry enough unique information to enable inversion~\cite{dusmanu2021privacy,pittaluga2023ldpfeat,pietrantoni2023segloc}. 
Obfuscation strength needs to be balanced with utility~\cite{pittaluga2023ldpfeat}, as the  obfuscated entities still need to be informative enough for  downstream tasks, \eg, feature matching and camera pose estimation for  localization. 

\phantomsection\label{sec:problem}%
Chelani \etal~\cite{chelani2021lineclouds,chelani2025obfuscation} showed that, by estimating which keypoints are neighbors from the descriptors alone~\cite{chelani2025obfuscation}, positions close to the original keypoints can be recovered for many geometric obfuscation schemes~\cite{speciale2019privacy,speciale2019queries,geppert2020privacy,geppert2021privacy,geppert2022partial,Pan_2023_ICCV,lee2023pairedpoint,moon2024rayclouds}, while some schemes are still believed to be safe~\cite{kim2026dcl,moon2026lattices}.

In this paper, we consider the limiting case of geometric obfuscation, where only descriptors are available, without keypoint coordinates.\footnote{In the context of visual localization, this limit case is not practical as some information about the keypoints needs to be sent to the server to facilitate camera pose estimation. For other applications with privacy implications, just sending descriptors might be practical.}
Our \emph{solver} is a set transformer that predicts one 2D position per input descriptor.
Using the recovered keypoint positions and the descriptors, the image content can then be recovered using existing image inversion methods.
We train multiple solvers on public images, one for each type of local feature. 

\phantomsection\label{sec:descriptorsoverlap}%
Our approach is based on a relatively simple key insight. 
Feature descriptors describe local patches around a keypoint. 
Thus, two descriptors with overlapping patches partially describe the same image content. 
This overlap in content can then be used to reason about the relative positioning of the two patches and thus the two keypoints. 

The relation between the descriptors and the overlap of their corresponding patches can be very complex if each patch is rotated and rescaled to its own keypoint's orientation and scale, as in SIFT~\cite{lowe2004sift}. 
However, many modern learned extractors, including ALIKED~\cite{zhao2023aliked}, SuperPoint~\cite{detone2018superpoint}, XFeat~\cite{potje2024xfeat}, DISK~\cite{tyszkiewicz2020disk} and LoMa~\cite{nordstrom2026loma}, produce upright descriptors, \ie, their image regions are aligned with the image axes.
These extractors also omit explicit per-keypoint scale normalization.
In this case, the relationship becomes easier to learn. 
The fact that our solvers are able to recover the keypoint positions quite accurately from the descriptors shows that they are able to exploit this shared information between descriptors (Sec.~\ref{sec:mechanism}).

For the descriptors studied, our results show that geometric obfuscation can be circumvented by ignoring the obfuscated keypoints and recovering their positions with our solvers. 
Naturally, this raises questions about privacy. 
Since there is no clear definition of privacy~\cite{pietrantoni2026vulnerability}, we use the following approach: 
We compare the images recovered using the keypoint positions predicted by our solvers to the images recovered using the ground truth keypoint positions. 
If the latter are privacy-sensitive, \ie, they might contain content that a user might not want to share, then being able to accurately reconstruct them shows that there is a potential privacy leak. 
Furthermore, we show qualitatively that recognizable objects, faces and some document text can be recovered. 

\customparagraph{Contributions.}
\textbf{(1)}~We show that upright local descriptors can reveal the relative 2D keypoint positions and recognizable image content even when all positions are discarded.
\textbf{(2)}~We explain how overlapping image regions in the same orientation and scale provide clues about relative keypoint positions, supported by controlled tests.
We also recover positions and reconstruct image content from oriented SIFT~\cite{lowe2004sift} when each descriptor is transmitted with its keypoint's scale and orientation, computed by the extractor (Secs.~\ref{sec:mechanism} and~\ref{sec:exp-sift}).
\textbf{(3)}~We propose a set transformer-based solver that predicts one 2D position per descriptor in an unordered set, up to a common translation.
We recover keypoint positions for five learned feature extractors, and the trained ALIKED solver also works on image domains unseen during training, including faces and documents (Sec.~\ref{sec:exp-recovery}).
\textbf{(4)}~Solvers adapted to subsampled or quantized descriptors still recover keypoint positions in several tested settings; stronger modifications also reduce localization accuracy (Sec.~\ref{sec:defenses}). 
\textbf{(5)}~We will release our source code and our trained models.

\begin{figure*}[t]
  \centering
  \includegraphics[width=\linewidth]{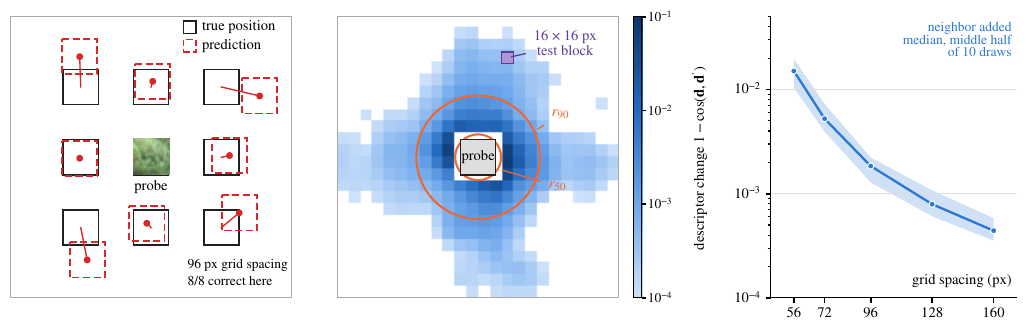}
  \caption{\textbf{ALIKED descriptor influence and position recovery.}
    A probe photograph stays fixed on a white background.
    \textbf{Left:} one neighbor photograph is added per trial; the solver uses only descriptors from both photographs.
    \swatchsolid~True and \swatchdashed~predicted positions have the correct direction in all eight trials shown.
    \textbf{Middle:} \swatchmap~descriptor change when a $16{\times}16$\,px block is added at each location, averaged over $15$ probe keypoints.
    \swatchring~$r_{50}$ and $r_{90}$ are median radii enclosing $50\%$ and $90\%$ of change in individual keypoint maps.
    \textbf{Right:} change in the probe's descriptors versus neighbor grid spacing.}
  \label{fig:mechanism-method}
\end{figure*}

\section{Related Work}
\label{sec:related}

\customparagraph{Inverting local features.}
Prior work reconstructs images from local descriptors such as SIFT and HOG, and from spatial CNN feature maps~\cite{weinzaepfel2011reconstructing,vondrick2013hoggles,mahendran2015understanding,dosovitskiy2016inverting}.
Images can also be reconstructed from sparse structure-from-motion points and their descriptors~\cite{pittaluga2019revealing,pietrantoni2026vulnerability}, and from colored point clouds without descriptors~\cite{song2020pointclouds}.
These methods use known spatial information.
Kato and Harada~\cite{kato2014reconstruction} discard feature positions but synthesize a texture rather than aiming to recover the original keypoint positions. 
We estimate 2D keypoint positions from descriptors alone, enabling image inversion. 

\customparagraph{Image and descriptor obfuscation.} 
Descriptor obfuscation approaches introduce ambiguities into the descriptors, \eg, by lifting descriptors to represent subspaces~\cite{dusmanu2021privacy}, through adversarial training designed to make image inversion quality worse~\cite{ng2022ninjadesc}, or through quantization and representing each feature through multiple descriptors~\cite{pittaluga2023ldpfeat}. 
Note that~\cite{dusmanu2021privacy,ng2022ninjadesc} still allow image inversion~\cite{pittaluga2023ldpfeat,pietrantoni2026vulnerability}, while~\cite{pittaluga2023ldpfeat} does not make any guarantees that image content cannot be recovered from their obfuscated descriptors and keypoints.
Recovering this geometry requires descriptors that are informative enough to reveal patch overlap. 
Recovery images from these obfuscated descriptor schemes is outside the scope of this work.
In this paper, we focus on obfuscating keypoint positions, \ie, the geometry of the features. 

Obfuscating the whole image~\cite{pietrantoni2023segloc,pietrantoni2025gsff,panek2026structureless} or only privacy-relevant parts~\cite{panek2026structureless,dangwal2021mitigating,vultaggio2026misanthropeprivacypreservingkeypointdetector} are alternatives to descriptor obfuscation. 
\Eg,~\cite{pietrantoni2023segloc,pietrantoni2025gsff,panek2026structureless} introduce ambiguity by replacing color values through segmentation labels, since this many-to-one mapping makes it harder to recover the original image content~\cite{pietrantoni2026vulnerability}. 
Panek \etal and Dangwal \etal~\cite{panek2026structureless,dangwal2021mitigating} identify potentially sensitive regions, such as text or faces, and either mask out the regions~\cite{panek2026structureless} or ensure that no keypoints are detected in these regions~\cite{dangwal2021mitigating,vultaggio2026misanthropeprivacypreservingkeypointdetector}. 
This line of work is not affected by our approach.

\customparagraph{Geometric obfuscation.} 
Orthogonal to descriptor obfuscation, one can also obfuscate the keypoint positions. 
Geometric obfuscation schemes typically introduce ambiguities by lifting keypoints to 2D lines~\cite{speciale2019queries,geppert2020privacy,geppert2021privacy,Pan_2023_ICCV,lee2023pairedpoint,kim2026dcl}. 
This leaves one or more degrees of freedom in the position of each keypoint. 
As a result, camera pose estimation becomes more complicated, and potentially less accurate, compared to using 2D keypoint positions. 
We study the limit of geometric obfuscation for image queries by discarding every keypoint position.
For the descriptors studied, we show that keypoint positions can be recovered from the descriptors alone, regardless of the geometric obfuscation scheme. 
Geometric obfuscation thus needs to be combined with techniques that make keypoint recovery harder, \eg, sparsifying the keypoints or using less informative descriptors; adapted solvers still recover positions in several such settings (Sec.~\ref{sec:defenses}). 

Related schemes obfuscate 3D maps~\cite{speciale2019privacy,geppert2022partial,Pan_2023_ICCV,lee2023pairedpoint,moon2024rayclouds}.
Scene information can be recovered from obfuscated point clouds~\cite{chelani2021lineclouds,chelani2025obfuscation}, camera poses~\cite{chelani2023poses}, or scene coordinate regression weights~\cite{nasypanyi2026weights}.

\customparagraph{Recovering obfuscated geometry.} 
Chelani \etal~\cite{chelani2021lineclouds,chelani2025obfuscation} recover approximate keypoint positions from most geometric obfuscation schemes~\cite{speciale2019privacy,speciale2019queries,geppert2022partial,Pan_2023_ICCV,lee2023pairedpoint} using which keypoints were nearest neighbors, information they extract from the descriptors~\cite{chelani2025obfuscation}. 
However, their approach needs to be adapted to each obfuscation scheme, and neighborhood information does not help in all cases~\cite{kim2026dcl,moon2026lattices}. 
Our solver instead recovers keypoint positions from descriptors, enabling image reconstruction without the obfuscated geometry.

%% file: tex/method.tex
\section{Method}
\label{sec:method}

Our goal is to estimate 2D keypoint positions from descriptors alone, then use those positions for image reconstruction (Fig.~\ref{fig:teaser}).
The client extracts local features from a private query image $I$ and transmits only the unordered descriptor set $\Descr=\{\descr_i\}_{i=1}^{\numkp}$, where $\descr_i\in\Real^{\dimd}$.
The corresponding positions $\Pos=\{\pos_i\in\Real^2\}_{i=1}^{\numkp}$ are withheld.
The recovered positions are defined up to a common translation.

We assume a known feature extractor, preprocessing and input image resolution, and train a separate solver for each extractor.
Our main setting uses upright learned descriptors, such as ALIKED, SuperPoint, XFeat, DISK and LoMa~\cite{zhao2023aliked,detone2018superpoint,potje2024xfeat,tyszkiewicz2020disk,nordstrom2026loma}, computed without per-keypoint rotation or scale normalization.
We recover positions in pixels of the input image.
SIFT descriptors accompanied by keypoint scale and orientation are evaluated as a separate input setting (Sec.~\ref{sec:exp-sift}).

The pipeline has two stages.
A set transformer, the \emph{solver}, predicts a position for each descriptor.
An image inversion network, the \emph{inverter}, reconstructs an image from the descriptors at those positions.
During training, keypoint positions extracted from public images supervise the solver and provide the inverter's spatial input.
For a private query, the solver receives only descriptors, and the inverter uses the predicted positions.

\subsection{The global set transformer solver}
\label{sec:solver}

\begin{figure}[t]
  \centering
  \includegraphics[width=1.0\linewidth]{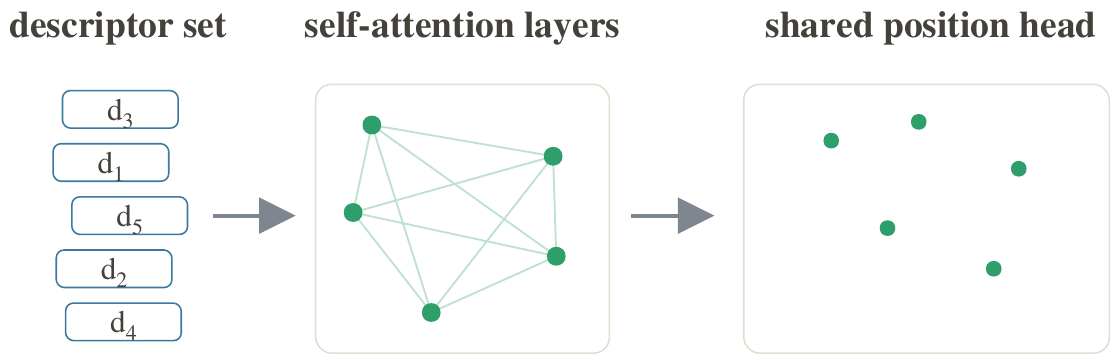}
  \caption{\textbf{From unordered descriptors to keypoint positions.}
    Each descriptor becomes a token, without positional encoding.
    Global self-attention combines the tokens; a shared head predicts one 2D position per descriptor in one forward pass.}
  \label{fig:arch}
\end{figure}

\customparagraph{Why a global solver?}
Nearby descriptors depend on overlapping image regions. In the common upright frame, shared content provides clues to the relative offset $\off_{ij}=\pos_j-\pos_i$ between keypoints.
Fig.~\ref{fig:mechanism-method} (left) tests position recovery from the two photographs' descriptors.
The extractor processes the whole image, so a blank gap between photographs does not isolate their descriptors.

The middle panel instead measures descriptor influence by adding a $16{\times}16$\,px textured block at one grid location at a time.
We measure change as $1-\cos(\descr,\descr')$ at fixed probe keypoints, without ($\descr$) and with ($\descr'$) the added content.
Repeating the neighbor-photograph experiment at different spacings (right) shows that its influence weakens with distance.
In this example, the farther diagonal neighbors have weaker influence and larger placement errors.
Joint prediction lets the solver combine weak responses from many descriptors.

\customparagraph{Architecture.}
The solver $\solver\colon\Descr\mapsto\{\posrec_i\in\Real^2\}_{i=1}^{\numkp}$ predicts all positions jointly, with learned network parameters~$\boldsymbol{\theta}$ (Fig.~\ref{fig:arch}).
It outputs one 2D position per descriptor and, optionally, a per-axis uncertainty estimate $\boldsymbol{\sigma}_i$ from a separate head.
A shared linear embedding converts each descriptor into one token, without positional encoding.
Self-attention lets every token use information from the whole set, and a shared output head predicts its position in one forward pass~\cite{zaheer2017deepsets,lee2019settransformer,vaswani2017attention}.
The solver is permutation-equivariant: reordering the descriptors reorders the predicted positions in the same way, preserving the descriptor-to-position correspondence.
No per-query optimization is performed.

\customparagraph{Training loss.}
We compare predicted and ground-truth (GT) positions after centering each set independently.
The centroid $\boldsymbol{\mu}(\Pos)=\frac{1}{\numkp}\sum_i\pos_i$ is the mean keypoint position.
The centered positions are $\Center(\Pos)=\Pos-\boldsymbol{\mu}(\Pos)$, obtained by subtracting this mean from every position.
We define the position error as $\boldsymbol{\Delta}=\Center(\Posrec)-\Center(\Pos)$ and train the position output with a smooth-$L_1$ (Huber) loss.
An optional uncertainty head predicts per-axis standard deviations $\boldsymbol{\sigma}_i$ and is trained with a Gaussian negative log-likelihood on the detached position error.
The combined loss is
\begingroup
\setlength{\abovedisplayskip}{12pt}
\setlength{\belowdisplayskip}{12pt}
\setlength{\abovedisplayshortskip}{12pt}
\setlength{\belowdisplayshortskip}{12pt}
\begin{equation}
  \label{eq:loss}
  \mathcal{L}=\mathrm{Huber}(\boldsymbol{\Delta})+\lambda\,\mathrm{NLL}(\mathrm{sg}[\boldsymbol{\Delta}];\boldsymbol{\sigma}),
\end{equation}
\endgroup
where $\lambda=1$ and $\mathrm{sg}$ stops gradients through $\boldsymbol{\Delta}$.
Details of detachment and shared-backbone training are in \suppref{sec:supp-impl}.

\subsection{From keypoint positions to an image}
\label{sec:inverter}

We place each descriptor on a feature grid as a small Gaussian centered at its predicted position.
A U-Net decodes that grid into an image~\cite{ronneberger2015unet}.
It is trained with GT positions from public images and uses predicted positions for query reconstruction.
Pixel metrics and all reconstruction examples shift the predicted positions to the GT centroid (Sec.~\ref{sec:exp-inversion}), except the teaser (Fig.~\ref{fig:teaser}), which places them at the image center, as an attacker without ground truth would.
The recovery and inversion networks are trained separately.
Other inverters accepting descriptors and positions can use the same predicted coordinates, but reconstruction quality depends on both the inverter and the position errors.
Section~\ref{sec:exp-inversion} evaluates these effects with GT-position references and inversion controls.

%% file: tex/experiments.tex
\section{Experiments}
\label{sec:experiments}

We first evaluate layout recovery across extractors and image domains and examine where the positional information comes from (Sec.~\ref{sec:exp-recovery}).
We then compare the image content reconstructed from recovered and from ground-truth (GT) positions (Sec.~\ref{sec:exp-inversion}) and extend the attack to SIFT released with scale and orientation (Sec.~\ref{sec:exp-sift}).
Finally, we test whether fewer or coarser descriptors prevent recovery while preserving localization accuracy (Sec.~\ref{sec:defenses}).

\phantomsection\label{sec:setup}%
\customparagraph{Setup.}
We refer to the assignment of a 2D position to each descriptor as the keypoint layout, defined up to a global translation. 
The solver is our set transformer (Sec.~\ref{sec:solver}), trained separately for each extractor.
We train on public COCO images~\cite{lin2014coco} and evaluate on a held-out split.

Our main extractor is ALIKED~\cite{zhao2023aliked}, with its upright, fixed-scale \texttt{aliked-n16} descriptors.
We use the authors' pretrained weights with detection threshold $0.2$ and keep at most $2048$ keypoints per image. 
Images are resized to $512$\,px on the longest side.
ALIKED upsamples them internally to $1024$\,px during extraction and maps the keypoint positions back to the $512$\,px frame.
We also evaluate four popular upright extractors, XFeat, SuperPoint, DISK and LoMa (Tab.~\ref{tab:main}; implementation details in \suppref{sec:supp-impl}).

\customparagraph{Metrics.}
\phantomsection\label{sec:metrics}%
Each layout metric is computed per image; we report its median over test images.
The \emph{residual}, or position error, is the median distance in pixels between recovered and GT keypoint positions after aligning the layout centroids.
The \emph{spacing} is the median distance from a GT keypoint to its nearest neighbor.
\emph{Res./Sp.} divides the median residual by the median spacing; at $1$, keypoints are typically misplaced by about one keypoint spacing.
\emph{dcorr} is the Pearson correlation between recovered and GT distances over all keypoint pairs.
\emph{kNN} is the fraction of each keypoint's five nearest neighbors in the GT layout that are also among its five nearest neighbors in the recovered layout, averaged over keypoints.
Both (dcorr, kNN) are $1$ for a perfect layout and ${\approx}0$ for random positions.

For image recovery, we use the reconstruction from GT keypoint positions as the \emph{reference reconstruction}, since this represents what an inverter can recover when the original features are available.
We compare reconstructions from recovered and random positions with this reference using PSNR and SSIM on color images, LPIPS~\cite{zhang2018lpips} on grayscale images, and object recovery (\emph{OR}).
OR measures the fraction of objects identified by Qwen2-VL~\cite{wang2024qwen2vl} in the reference that are also identified in the reconstruction.
To measure the inverter’s reconstruction error, we compare the original image with the image reconstructed using the GT keypoint positions.

\subsection{Layout recovery}
\label{sec:exp-recovery}

\begin{table}
        \caption{\textbf{Layout recovery across extractors and image domains.}
      Top: $200$ held-out COCO images~\cite{lin2014coco}. Bottom: five other domains.
      Residual: median ($25$th--$75$th percentile) over images; Res./Sp.: residual divided by spacing (Sec.~\ref{sec:metrics}).
      Recovered positions assigned to keypoints at random: residual ${\approx}190$--$230$\,px, kNN and dcorr $0.00$ (COCO).
      In training and at test time, at most $2048$ keypoints per image for every extractor.}
  \centering
  \footnotesize
  \setlength{\tabcolsep}{2pt}
  \begin{tabular*}{\linewidth}{@{\extracolsep{\fill}}lccccc@{}}
    \toprule
    & \textbf{Spacing} & \textbf{Residual (px)} $\downarrow$ & \textbf{Res./Sp.} & \textbf{kNN} $\uparrow$ & \textbf{dcorr} $\uparrow$ \\
    \midrule
    \multicolumn{6}{@{}l}{\textit{COCO, each extractor with its own solver}} \\
    ALIKED~\cite{zhao2023aliked} & 4.2 & 4.0 {\scriptsize(3.3--5.1)} & 0.94 & 0.89 & 0.999 \\
    XFeat~\cite{potje2024xfeat} & 5.1 & 2.9 {\scriptsize(2.6--3.4)} & 0.57 & 0.91 & 1.000 \\
    SuperPoint~\cite{detone2018superpoint} & 3.0 & 4.7 {\scriptsize(4.1--5.5)} & 1.57 & 0.88 & 0.998 \\
    DISK~\cite{tyszkiewicz2020disk} & 5.7 & 3.2 {\scriptsize(2.8--3.8)} & 0.57 & 0.87 & 0.999 \\
    LoMa~\cite{nordstrom2026loma} & 3.5 & 7.0 {\scriptsize(5.7--9.3)} & 1.99 & 0.83 & 0.997 \\
    \specialrule{\lightrulewidth}{\aboverulesep}{1pt}
    \specialrule{\lightrulewidth}{0pt}{\belowrulesep}
    \multicolumn{6}{@{}l}{\textit{ALIKED solver trained on COCO, out-of-domain, no retraining}} \\
    MegaDepth~\cite{li2018megadepth} & 4.1 & 4.1 {\scriptsize(3.4--4.9)} & 0.98 & 0.90 & 0.999 \\
    DIV2K~\cite{agustsson2017div2k} & 4.3 & 4.3 {\scriptsize(3.3--5.7)} & 1.00 & 0.89 & 0.999 \\
    ADE20K~\cite{zhou2017ade20k} & 4.4 & 4.2 {\scriptsize(3.4--5.3)} & 0.95 & 0.90 & 0.999 \\
    CelebA-HQ~\cite{karras2018celebahq} & 4.5 & 5.6 {\scriptsize(4.4--6.8)} & 1.25 & 0.88 & 0.999 \\
    FUNSD forms~\cite{jaume2019funsd} & 3.4 & 5.9 {\scriptsize(4.6--7.4)} & 1.73 & 0.88 & 0.998 \\
    \bottomrule
  \end{tabular*}
  \label{tab:main}\label{tab:ood}
\end{table}

\phantomsection\label{sec:exp-ood}%
\customparagraph{Recovery across extractors and domains.}
Using only unordered descriptors, the solver recovers the keypoint layouts of all five extractors with a median position error of $0.6$--$2.0$ times the median nearest-neighbor distance between GT keypoints (Res./Sp., Tab.~\ref{tab:main}).
For ALIKED, the residual is $4.0$\,px against a $4.2$\,px spacing, and on average $4.5$ of each keypoint's five nearest neighbors are kept (kNN $0.89$).
Even LoMa, whose upright descriptors are partly robust to rotation, reveals its layout.
The ALIKED solver also generalizes without retraining from COCO to five public image collections:
$806$ landmark photographs (all images of the two MegaDepth-1500 test scenes~\cite{li2018megadepth}),
$300$ high-resolution photographs (DIV2K training set~\cite{agustsson2017div2k}),
$500$ indoor and outdoor scenes (ADE20K validation set~\cite{zhou2017ade20k}),
$200$ faces (CelebA-HQ~\cite{karras2018celebahq}) and the $50$ FUNSD test forms~\cite{jaume2019funsd}.
Across these domains, the residual stays between $4.1$ and $5.9$\,px (Tab.~\ref{tab:main} Bottom, Fig.~\ref{fig:inversion}).

We compare our set transformer with four alternatives (Tab.~\ref{tab:solvers}).
Classical MDS~\cite{torgerson1952mds} and Isomap~\cite{tenenbaum2000isomap} embed descriptors according to their distances in descriptor space, but recover little of the layout.
Deep Sets~\cite{zaheer2017deepsets} processes each descriptor independently and combines it with a pooled set representation. It recovers the coarse arrangement but not the local neighborhoods.
A graph network~\cite{wang2019dgcnn} that exchanges information between similar descriptors performs better, while our set transformer, which allows every descriptor to attend to all others, is the most accurate.
These results show that accurate layout recovery benefits from combining information across the descriptor set.

\begin{table}
  \caption{\textbf{Comparing methods for recovering the keypoint layout.}
    All methods use ALIKED descriptors and $200$ held-out COCO images~\cite{lin2014coco}; the learned ones are trained on the same $15$k images for $80$ epochs with about $14.2$M parameters each.
    MDS and Isomap need no training and are aligned to the ground truth by rotation, reflection, scale and translation, the learned methods by translation only.
    Our set transformer is the most accurate; with the full training set it reaches $4.0$\,px residual error (Tab.~\ref{tab:main}).}
  \centering
  \footnotesize
  \setlength{\tabcolsep}{2pt}
  \begin{tabular*}{\linewidth}{@{\extracolsep{\fill}}lccc@{}}
    \toprule
    \textbf{Solver} & \textbf{Residual (px)} $\downarrow$ & \textbf{dcorr} $\uparrow$ & \textbf{kNN} $\uparrow$ \\
    \midrule
    Classical MDS~\cite{torgerson1952mds}   & 137.9 & 0.04 & 0.04 \\
    Isomap~\cite{tenenbaum2000isomap}       & 124.3 & 0.18 & 0.09 \\
    Deep Sets~\cite{zaheer2017deepsets}     & 28.6  & 0.93 & 0.27 \\
    Descriptor-kNN GNN~\cite{wang2019dgcnn} & 11.5  & 0.99 & 0.69 \\
    \textbf{Set transformer (ours)}         & \textbf{6.1} & \textbf{1.00} & \textbf{0.84} \\
    \bottomrule
  \end{tabular*}
  \label{tab:solvers}
\end{table}

\begin{figure*}[t]
  \centering
  \includegraphics[width=0.99\linewidth]{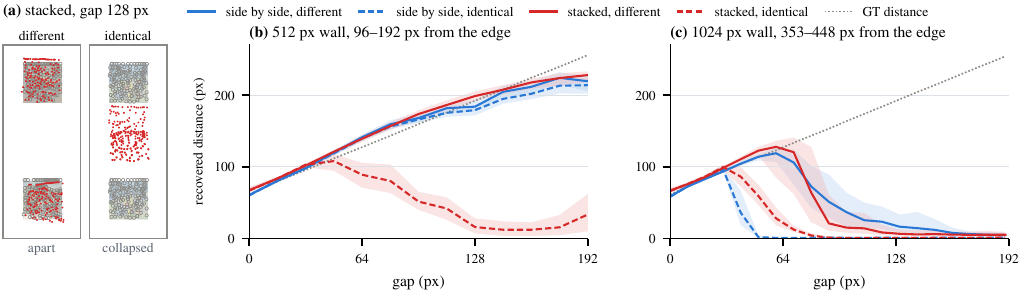}
    \caption{\textbf{Far from the image border, two paintings stay apart only at small gaps.}
    Two $64$\,px paintings~\cite{nga2026openaccess} on a plain wall, $100$ pairs per condition.
    (a)~Stacked pairs at a gap of $128$\,px on the $512$\,px wall: different paintings stay apart, identical ones collapse; gray circles: GT keypoints, red dots: recovered keypoints.
    (b,c)~Recovered distance against the gap on the $512$\,px wall, $96$--$192$\,px from the nearest image edge (b), and on a $1024$\,px wall, $353$--$448$\,px from it (c), for different (solid) and identical (dashed) paintings, side by side (blue) or stacked (red); dotted gray: GT distance.
    Medians over pairs with interquartile bands.}
  \label{fig:paintings}
\end{figure*}

\phantomsection\label{sec:mechanism}%
\customparagraph{When does relative-position recovery work?}
We place two different or identical $64$\,px paintings~\cite{nga2026openaccess} side by side or stacked on a plain wall, vary the gap between their edges and measure the distance between the centroids of their keypoints as recovered by the COCO-trained ALIKED solver (Fig.~\ref{fig:paintings}).
On a $1024$\,px wall, at least $353$\,px from every edge, only the paintings themselves tell the solver where they lie relative to each other (Fig.~\ref{fig:paintings}c).
This wall is larger than any training image we used, but small gaps are still recovered accurately.
Different paintings keep their distance up to gaps of about $50$\,px and nearly coincide from about $100$\,px.
While the gap is small, descriptors near it also capture part of the other painting, whose influence fades with distance (Fig.~\ref{fig:mechanism-method}, right), and the solver uses this shared content to place the two paintings (\suppref{sec:supp-copies}).
In real images, intervening keypoints can form chains between regions (\suppref{sec:supp-connectivity}).
Identical copies, with nearly identical descriptors, lose their distance from $40$--$50$\,px.
On a $512$\,px wall, as in our test images, the descriptors also change with where a painting lies in the image (Fig.~\ref{fig:paintings}b).
This position cue keeps different paintings, and identical ones side by side, apart up to gaps of $160$\,px, but stacked identical copies collapse from about $64$\,px.
The two copies differ only in where they lie in the image, and ALIKED's descriptors change less with vertical than with horizontal position.
At a $128$\,px gap, stacked copies are therefore more alike than side-by-side ones (median cosine distance between their descriptors at the same keypoints $0.013$ vs.\ $0.018$), and the solver also needs a larger difference to separate copies vertically (\suppref{sec:supp-copies}).

\subsection{Image inversion}
\label{sec:exp-inversion}

\begin{table}
\caption{\textbf{Image reconstructions from recovered and GT positions are similar.}
  Results on the $200$ held-out COCO images~\cite{lin2014coco}.
  Recovered and random positions are compared against reference reconstructions from GT positions, using the same descriptors and inverter.
  OR measures the fraction of reference objects recovered.
  Italic rows compare the reference reconstructions against the original images instead, providing a measure of inversion quality.
  Other domains:~\suppref{sec:supp-oodres}; every metric of the benchmark:~\suppref{sec:supp-aachen}.}
  \centering
  \footnotesize
  \setlength{\tabcolsep}{2pt}
  \begin{tabular*}{\linewidth}{@{\extracolsep{\fill}}llcccc@{}}
    \toprule
    \textbf{Inverter} & \textbf{Positions} & \textbf{PSNR} $\uparrow$ & \textbf{SSIM} $\uparrow$ & \textbf{LPIPS} $\downarrow$ & \textbf{OR} $\uparrow$ \\
    \midrule
    U-Net (ours) & recovered & 21.4 & 0.68 & 0.10 & 0.63 \\
                 & random    & 12.1 & 0.23 & 0.60 & 0.07 \\
                 & \textit{GT vs.\ original} & \textit{18.2} & \textit{0.69} & \textit{0.23} & \textit{0.54} \\
    \midrule
    Diffusion~\cite{pietrantoni2026vulnerability} & recovered & 16.4 & 0.50 & 0.24 & 0.68 \\
                 & random    & 11.6 & 0.18 & 0.66 & 0.07 \\
                 & \textit{GT vs.\ original} & \textit{16.1} & \textit{0.50} & \textit{0.28} & \textit{0.65} \\

    \bottomrule
  \end{tabular*}
  \label{tab:image}
\end{table}

\begin{figure}[t]
  \centering
  \includegraphics[width=0.8\linewidth]{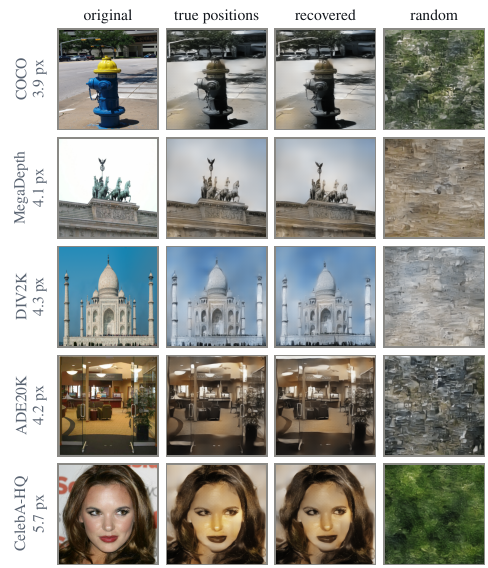}
\caption{\textbf{Recovered keypoint positions reveal recognizable image content across domains.}
    Examples from COCO and four unseen domains (Tab.~\ref{tab:main}), selected near each domain's median residual.
    \textbf{True positions}: reference reconstruction, from the GT keypoint positions.
    \textbf{Recovered}: reconstruction using positions predicted from unordered descriptors.
    \textbf{Random}: reconstruction using the same descriptors at random positions.
    All three use the same U-Net inverter.}
  \label{fig:inversion}
\end{figure}

\customparagraph{Recovered positions preserve image content.}
We compare images reconstructed from recovered positions with those reconstructed from the GT positions, using the same descriptors and inverter (Tab.~\ref{tab:image}, Fig.~\ref{fig:inversion}).
With our U-Net, the two reconstructions are close (LPIPS $0.10$), and $63\%$ of the objects identified in the reference reconstruction are also found in the recovered one.
Placing the same descriptors at random positions instead reduces object recovery to $7\%$.
Thus, recovering the descriptor-to-position correspondence is essential for reconstructing the scene.

We also evaluate the diffusion inverter of~\cite{pietrantoni2026vulnerability}.
It recovers a similar fraction of objects ($68\%$), although its reconstructions differ more from the reference reconstruction (LPIPS $0.24$).
Part of this difference comes from stochastic sampling: two reconstructions from the same positions already differ by LPIPS $0.18$.
Figures use the deterministic U-Net, avoiding this variation and relying less on a generative prior.

\customparagraph{Faces retain their identity.}
On CelebA-HQ~\cite{karras2018celebahq}, ArcFace~\cite{deng2019arcface} detects a face in $198$ of $200$ reconstructions from recovered positions and identifies $98\%$ of them among the $200$ originals, against $100\%$ from GT positions (Fig.~\ref{fig:inversion}, \suppref{sec:supp-oodres}).
With random positions, no face is detected.
The recovered layout thus preserves facial identity.

\begin{figure}[t]
  \centering
  \includegraphics[width=\linewidth]{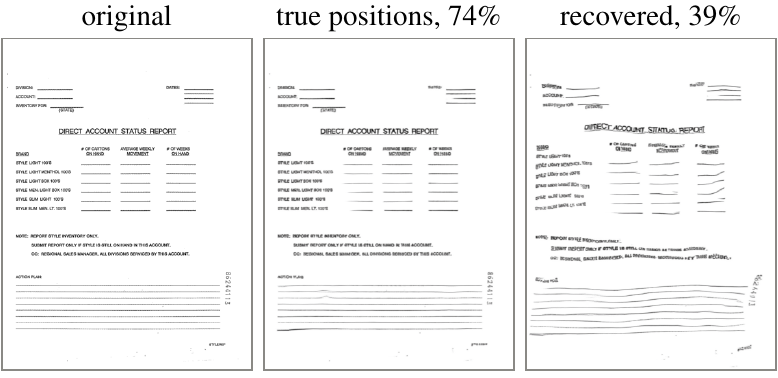}
  \caption{\textbf{Reconstructing a document form.}
    A FUNSD~\cite{jaume2019funsd} form and its reconstructions from GT and recovered positions, by a document-trained U-Net using the same descriptors.
    Percentages: share of the form's words read by Qwen2-VL~\cite{wang2024qwen2vl} in each image; medians over all $50$ forms: $50\%$ from GT and $15\%$ from recovered positions (Tab.~\ref{tab:ocr}).}
  \label{fig:documents}
\end{figure}

\customparagraph{Text is harder to recover.}
The solver also recovers the layout of document images without retraining (Tab.~\ref{tab:main}), but recovering readable text requires greater positional accuracy.
We train a separate U-Net on RVL-CDIP~\cite{harley2015rvlcdip} documents and evaluate it on $50$ unseen FUNSD~\cite{jaume2019funsd} forms.
We give the full reconstructed form to Qwen2-VL~\cite{wang2024qwen2vl} and GPT-5.5~\cite{OpenAI2026GPT55} and ask each model to transcribe all visible text, without cropping individual text regions (Tab.~\ref{tab:ocr}, Fig.~\ref{fig:documents}).
From the reference reconstructions they transcribe about half of the original words.
From recovered positions, their recall drops to $15\%$ and $2\%$, respectively, with low precision.
The recovered images preserve the document structure and some larger text, but most words remain unreadable.
Even moving keypoints independently by about $2$\,px already leaves no word readable (\suppref{sec:supp-precision}), and the recovered FUNSD layouts have a median residual of $5.9$\,px.

\begin{table}
  \caption{\textbf{Most text is lost when reconstructing from recovered positions.}
      Qwen2-VL~\cite{wang2024qwen2vl} and GPT-5.5~\cite{OpenAI2026GPT55} transcribe $50$ FUNSD~\cite{jaume2019funsd} forms, either directly from the original image or from our document-trained U-Net's reconstructions.
      Recall is the fraction of GT words recovered; precision is the fraction of transcribed words that appear on the form.
      Medians over forms.}
  \centering
  \footnotesize
  \setlength{\tabcolsep}{2pt}
  \begin{tabular*}{\linewidth}{@{\extracolsep{\fill}}lcccc@{}}
    \toprule
    & \multicolumn{2}{c}{\textbf{Qwen2-VL}} & \multicolumn{2}{c}{\textbf{GPT-5.5}} \\
    \cmidrule(lr){2-3}\cmidrule(lr){4-5}
    \textbf{Image} & \textbf{recall} & \textbf{precision} & \textbf{recall} & \textbf{precision} \\
    \midrule
    Original form               & 0.78 & 0.82 & 0.90 & 0.91 \\
    GT positions                & 0.50 & 0.58 & 0.51 & 0.68 \\
    Recovered positions         & 0.15 & 0.21 & 0.02 & 0.20 \\
    Random positions            & 0.02 & 0.10 & 0.00 & 0.00 \\
    \bottomrule
  \end{tabular*}
  \label{tab:ocr}
\end{table}

\subsection{SIFT with released scale and orientation}
\label{sec:exp-sift}

Oriented SIFT~\cite{lowe2004sift} computes descriptors in each keypoint's dominant orientation, unlike our upright setting (Sec.~\ref{sec:problem}).
Our solver does not reliably recover positions from these descriptors alone.
We therefore consider SIFT records released with keypoint scale and orientation, as in OpenCV~\cite{opencv_library} and COLMAP~\cite{schoenberger2016mvs,schoenberger2016sfm}, while positions remain withheld. Further details in \suppref{sec:supp-sift}.

\begin{figure}[t]
  \centering
  \includegraphics[width=0.8\linewidth]{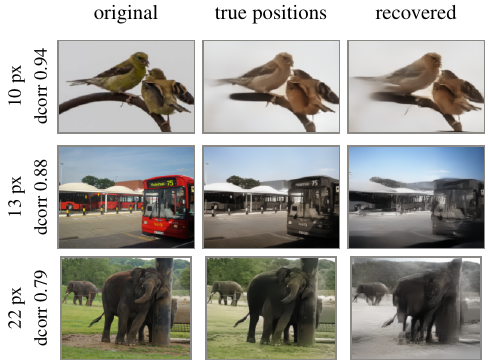}
  \caption{\textbf{Oriented SIFT with scale and orientation.}
    Two well-recovered scenes and a median one; position error and dcorr on the left.
    The same inverter uses upright descriptors computed from the image at GT positions, or MLP estimates at recovered positions.}
  \label{fig:sift-released-metadata-render}
\end{figure}

A small MLP, trained on public images, first estimates each upright descriptor from its oriented descriptor and orientation.
The solver then predicts positions from these estimates, with scale as an additional input.
An inverter trained on upright SIFT reconstructs images from the estimated descriptors at the recovered positions (Fig.~\ref{fig:sift-released-metadata-render}).
The median position error over test images is $22$\,px (dcorr $0.82$), about five keypoint spacings.
Recovery when orientation is withheld remains open.

%% file: tex/defenses.tex
\subsection{Ablating Robustness}
\label{sec:defenses}

\begin{figure*}[!t]
  \centering
  \includegraphics[width=1.0\linewidth]{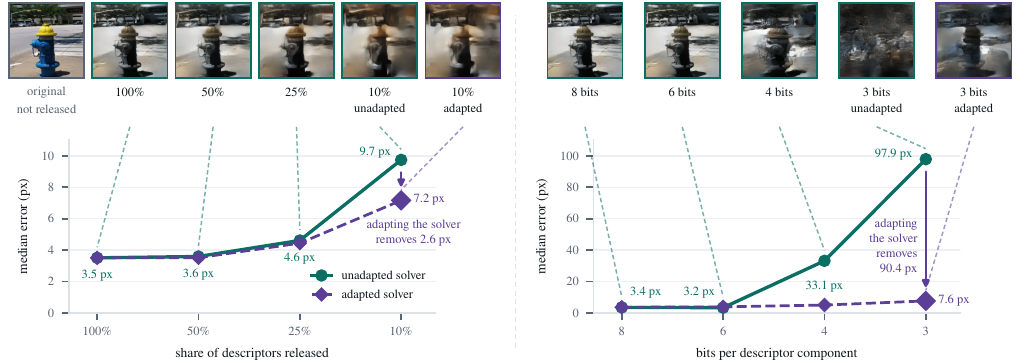}%
  \begin{subfigure}{0pt}\phantomcaption\label{fig:subsample-inversion}\end{subfigure}%
  \begin{subfigure}{0pt}\phantomcaption\label{fig:quantize-inversion}\end{subfigure}
  \caption{\textbf{An adapted solver recovers layouts from sparse or quantized descriptors.}
    Results on $200$ held-out COCO images, showing median position error.
    \textbf{Left}: a random fraction of each image's descriptors is retained.
    \textbf{Right}: each descriptor component is quantized to $2^b$ equally spaced values in $[-1,1]$ using $b$ bits, then the descriptor is renormalized.
    Unadapted and adapted solvers receive $10$ additional training epochs on unmodified and correspondingly modified descriptors, respectively.
    \textbf{Top}: U-Net reconstructions of one test image from recovered positions and the released descriptors.
    \textbf{Bottom}: layout errors for the two solvers.}
  \label{fig:defense-inversion}
\end{figure*}

We test whether releasing fewer or coarser descriptors reduces layout leakage, and we also measure the associated localization cost.
For each modification, we fine-tune an \emph{adapted} solver on descriptors modified in the same way, and compare it with an \emph{unadapted} solver trained for the same number of epochs on unmodified descriptors (Fig.~\ref{fig:defense-inversion}).
These steps follow the adaptive-evaluation practice~\cite{carlini2019evaluating,tramer2020adaptive}. 
On COCO, both start from the ALIKED solver of Sec.~\ref{sec:exp-recovery} and train for $10$ more epochs; on Aachen, both start from the LoMa solver and train for $35$ more epochs (\suppref{sec:supp-defenses}).

\customparagraph{Subsampling and quantization on COCO.}
We randomly retain $10$--$50\%$ of each image's descriptors or quantize their $128$ components to $3$--$8$ bits (Fig.~\ref{fig:defense-inversion}).
Both modifications degrade the unadapted solver, but adaptation reduces the error, most strongly under quantization (Fig.~\ref{fig:defense-inversion}).
The illustrated reconstructions also become more recognizable after adaptation (Fig.~\ref{fig:defense-inversion}).
We also test farthest-first spatial selection against matched-count random subsets (\suppref{sec:supp-defenses}).

\begin{table}[t]
  \caption{\textbf{Localization and image recovery from modified query descriptors on Aachen Day--Night~\cite{sattler2018benchmarking}.}
    B: bytes per descriptor.
    $N$ keypoints: the $N$ highest-scoring keypoints.
    ${\geq}D$\,px ($n$): keypoints taken in score order, each kept only if at least $D$\,px from those already kept; $n$ is the median count.
    OPQ $m{\times}b$: $m$ sub-vectors coded with $b$ bits each, so B${}=mb/8$.
    Loc.: queries localized within $25$\,cm and $2^\circ$ against a $4$-bit map.
    Error: median position residual (px) of the adapted solver; unadapted errors in \suppref{sec:supp-aachen}.
    OR vs.\ orig.: object recovery of the U-Net reconstructions from GT and from recovered (Rec.) positions against the original image.
    LPIPS vs.\ GT: between the reconstruction from recovered positions and the one from GT positions; low with few keypoints, where both are mostly empty.
    Random positions: all $5{,}000$ descriptors at uniformly random positions.}
  \label{tab:loc}
  \centering
  \footnotesize
  \setlength{\tabcolsep}{1.7pt}
    \begin{tabular*}{\linewidth}{@{\extracolsep{\fill}}lrrrrrrr@{}}
    \toprule
    & & \multicolumn{2}{c}{\textbf{Loc. (\%)} $\uparrow$}
    & \textbf{Error}
    & \multicolumn{2}{c}{\textbf{OR vs.\ orig.}}
    & \textbf{LPIPS} \\
    \cmidrule(lr){3-4}
    \cmidrule(lr){6-7}
    \textbf{Query release} & \textbf{B}
    & \textbf{Day} & \textbf{Night}
    & \textbf{(px)} $\downarrow$
    & \textbf{GT} & \textbf{Rec.}
    & \textbf{vs.\ GT} \\
    \midrule
    $5{,}000$ keypoints & 1024 & 88.2 & 90.8 & 8.9 & 0.57 & 0.56 & 0.16 \\
    $1{,}024$ keypoints & 1024 & 87.3 & 82.7 & 9.0 & 0.50 & 0.47 & 0.17 \\
    $512$ keypoints & 1024 & 86.0 & 79.6 & 10.6 & 0.38 & 0.37 & 0.20 \\
    $256$ keypoints & 1024 & 83.5 & 76.5 & 13.1 & 0.27 & 0.25 & 0.20 \\
    $128$ keypoints & 1024 & 78.2 & 73.5 & 16.2 & 0.21 & 0.18 & 0.18 \\
    $64$ keypoints & 1024 & 71.6 & 56.1 & 21.3 & 0.14 & 0.10 & 0.16 \\
    \midrule
    ${\geq}50$\,px ($119$) & 1024 & 61.0 & 43.9 & 13.7 & 0.16 & 0.17 & 0.18 \\
    ${\geq}100$\,px ($35$) & 1024 & 24.1 & 9.2 & 25.5 & 0.00 & 0.01 & 0.17 \\
    ${\geq}200$\,px ($11$) & 1024 & 2.2 & 0.0 & 50.6 & 0.00 & 0.01 & 0.10 \\
    \midrule
    OPQ $16{\times}8$ & 16 & 82.5 & 79.6 & 15.5 & 0.53 & 0.42 & 0.30 \\
    OPQ $8{\times}8$ & 8 & 61.2 & 37.8 & 27.7 & 0.49 & 0.34 & 0.43 \\
    OPQ $4{\times}8$ & 4 & 15.4 & 8.2 & 76.9 & 0.48 & 0.17 & 0.56 \\
    \midrule
    Random positions & 1024 & -- & -- & -- & 0.57 & 0.14 & 0.59 \\
    \bottomrule
  \end{tabular*}
  \vspace{-5pt}
\end{table}

\customparagraph{Localization cost on Aachen.}
COCO lacks the map and ground-truth poses needed to evaluate localization, so we use Aachen Day--Night~\cite{sattler2018benchmarking} (Tab.~\ref{tab:loc}).
We modify the $922$ query descriptor sets by retaining the highest-scoring keypoints, selecting spatially separated keypoints, or compressing descriptors with optimized product quantization (OPQ)~\cite{jegou2011product,ge2013optimized,douze2024faiss}.
More details are in \suppref{sec:supp-aachen}.

Reducing the number of keypoints gradually degrades localization (Tab.~\ref{tab:loc}).
Fewer keypoints make the reconstruction less recognizable already from GT positions, recovering $57\%$ of the original objects with all keypoints, $27\%$ with $256$ and $14\%$ with $64$, the level of random positions.
Hiding the positions barely changes this: from recovered positions, the reconstructions recover $56\%$, $25\%$ and $10\%$ and stay close to those from GT positions (LPIPS $0.16$--$0.20$).
Releasing fewer keypoints, not hiding their positions, thus reduces what the descriptors reveal; $64$ keypoints still localize $71.6\%$ of day queries.
Spaced keypoints localize worse without protecting more: $119$ keypoints at least $50$\,px apart localize fewer day queries than the $128$ highest-scoring ones ($61.0\%$ vs.\ $78.2\%$), and their reconstructions recover about as many objects ($17\%$ vs.\ $18\%$).
Quantization keeps the GT-position reconstructions recognizable (at least $48\%$ of the objects) but makes the layout harder to recover: with $16$-byte OPQ codes, localization stays above $79\%$ and the reconstructions from recovered positions recover $42\%$ of the objects; with $4$ bytes, only $17\%$, but localization falls to $15.4\%$.

\customparagraph{Distractor descriptors.}
We also test whether adding descriptors of unrelated images hides the layout.
It does not: the solver flags descriptors drawn from many other images, and descriptors of one to three whole other images raise the error of the image's own keypoints only to $11$--$21$\,px (\suppref{sec:supp-defenses}).

%% file: tex/conclusion.tex
\section{Conclusion}
\label{sec:conclusion}
Withholding keypoint positions is meant to prevent image inversion; however, the upright descriptors we tested still contain enough information to recover them.
Because neighboring descriptors describe overlapping image regions, a set transformer trained on public images can recover the keypoint layouts of five extractors from the descriptors alone, with median errors low enough to reconstruct recognizable images, even on image domains it was not trained~on.
Regarding oriented SIFT, recovery still works when scale and orientation are provided.

Sharing fewer or quantized descriptors does not make discarding the positions much better at preventing image inversion.
On Aachen, reconstructions from recovered positions reveal nearly as much as those from GT positions.
Quantizing the descriptors does widen this gap, but it also reduces localization accuracy.

Our results show that, for the descriptors studied here, discarding keypoint positions hides neither the keypoint position nor the recognizable image content. 
Evaluations of geometric obfuscation should compare what reconstructions from recovered and GT positions reveal, alongside localization accuracy.

%% file: tex/acknowledgments.tex
\section*{Acknowledgments}
This work was supported by 
the Czech Science Foundation (GACR) EXPRO (grant no. 23-07973X),
the Ministry of Education, Youth and Sports of the Czech Republic through the e-INFRA CZ (ID:90254).
The author(s) used generative AI to assist with editing the manuscript, cleaning up figures and coding. All was reviewed by the authors, who take full responsibility for its content.

%% file: tex/bibliography.tex
{
    \small
    \bibliographystyle{ieeenat_fullname}
    \bibliography{bib/abbreviation_short,bib/bibliography}
}

%% file: tex/suppl.tex
\clearpage
\maketitlesupplementary

\setcounter{section}{0}
\renewcommand{\thesection}{S\arabic{section}}
\renewcommand{\theHsection}{S\arabic{section}}
\setcounter{table}{0}\setcounter{figure}{0}
\renewcommand{\thetable}{S\arabic{table}}\renewcommand{\theHtable}{S\arabic{table}}
\renewcommand{\thefigure}{S\arabic{figure}}\renewcommand{\theHfigure}{S\arabic{figure}}

This supplement gives the details and further results behind the paper: the implementation
(Sec.~\ref{sec:supp-impl}), the datasets (Sec.~\ref{sec:supp-datasets}) and feature extractors
(Sec.~\ref{sec:supp-extractors}), the measurements behind the mechanism (Sec.~\ref{sec:supp-mechanism}), the
reference reconstructions and metrics (Secs.~\ref{sec:supp-reference} and~\ref{sec:supp-metrics}), recovery and
inversion across domains (Sec.~\ref{sec:supp-oodres}), SIFT with released scale and orientation
(Sec.~\ref{sec:supp-sift}), the defenses on COCO (Sec.~\ref{sec:supp-defenses}) and on Aachen Day--Night
(Sec.~\ref{sec:supp-aachen}), the two paintings (Sec.~\ref{sec:supp-copies}), documents
(Sec.~\ref{sec:supp-documents}) and layout precision (Sec.~\ref{sec:supp-precision}). Section, table and figure
numbers without an S refer to the paper. We will make our code and trained models publicly available.

\section{Implementation details}
\label{sec:supp-impl}\suppfrozen{1}

\customparagraph{Solver.}
The solver $\solver$ (Fig.~\ref{fig:arch}) is a set
transformer~\cite{lee2019settransformer,vaswani2017attention,zaheer2017deepsets}. A linear layer turns each
descriptor into a token, self-attention layers combine the tokens, and a shared head predicts one 2D position
per token. The network has no positional encoding, so permuting its input permutes its output.
Table~\ref{tab:supp-impl} lists the training settings.

\begin{table}[!h]
  \centering
  \footnotesize
  \setlength{\tabcolsep}{3pt}
  \begin{tabularx}{\linewidth}{@{}lXX@{}}
    \toprule
    & \textbf{Solver} & \textbf{Inverter} \\
    \midrule
    Network & set transformer, pre-LN & U-Net \\
    Size & width $384$, $8$ layers, $8$ heads & base width $64$, depth $4$ \\
    Input & $K{=}768$ descriptors & Gaussians ($\sigma{=}2$) on a $256{\times}256$ grid \\
    Loss & Eq.~\ref{eq:loss}, $\lambda{=}1$ & L1 $+\,0.1\,{\times}$ VGG perceptual \\
    Optimizer & Adam, lr $5{\times}10^{-4}$ & Adam, lr $2{\times}10^{-4}$ \\
    Epochs, batch & $80$, $8$ & $250$, $16$ \\
    Training data & COCO train2017 ($117$k) & COCO, GT positions \\
    \bottomrule
  \end{tabularx}
  \caption{\textbf{Training settings.} Both networks hold out $15\%$ of their training images for
    validation. The DISK and LoMa solvers add a $2$-epoch linear learning-rate warmup and gradient
    clipping at $1.0$.}
  \label{tab:supp-impl}
\end{table}

\customparagraph{Loss.}
The loss of Eq.~\ref{eq:loss} combines a Huber term on the centered position error $\boldsymbol{\Delta}$ with a
Gaussian negative log-likelihood for the uncertainty head, which predicts a log variance per axis. The likelihood
sees $\boldsymbol{\Delta}$ only through the stop-gradient $\mathrm{sg}$, which detaches it. This way the
uncertainty head learns how large the error is without rescaling the position gradients, which can hinder
training~\cite{seitzer2022pitfalls}. Both heads share the transformer backbone and are trained together with
this one loss.

\customparagraph{Extractors.}
ALIKED uses the \texttt{aliked-n16} model with the authors' weights and detection threshold $0.2$
(Sec.~\ref{sec:setup}), run through LightGlue~\cite{lindenberger2023lightglue} with its default preprocessing.
ALIKED and LoMa (the LoMa-B model) see the image resized to $512$\,px on the longer side. XFeat, SuperPoint and
DISK follow the extraction recipe of the inversion benchmark~\cite{pietrantoni2026vulnerability} and see a
$512{\times}512$ square. Positions are measured in the frame each extractor sees. On COCO, every extractor keeps
its $2048$ highest-scoring keypoints.

\customparagraph{Inverter.}
As in Sec.~\ref{sec:inverter}, each descriptor is placed on a feature grid as a small Gaussian centered at its
position, and a U-Net~\cite{ronneberger2015unet} decodes that grid into an image. It is trained on ground-truth (GT)
positions and applied to recovered ones, so its output reflects both its own quality and the error of the
recovered layout.

\customparagraph{Solvers of Table~\ref{tab:solvers}.}
All learned solvers are trained with our settings on the same images and with the same seed, without tuning per
architecture. The GNN runs EdgeConv~\cite{wang2019dgcnn} on a graph that links each descriptor to its nearest
neighbors in descriptor space, so it never sees a position. Classical MDS~\cite{torgerson1952mds} and
Isomap~\cite{tenenbaum2000isomap} embed the cosine distances between the descriptors, Isomap through geodesics
on the same kind of graph. Their output has an arbitrary rotation, reflection and scale, so we align it to the
ground truth with a similarity transform before measuring the residual. This alignment pulls any uninformative
layout toward the centroid, so even a random layout gets a moderate residual ($142$\,px). For these two
methods we report dcorr and kNN evaluation values.

\section{Datasets}
\label{sec:supp-datasets}\suppfrozen{2}

\customparagraph{Training.}
The solvers and our U-Net are trained on public COCO train2017 images~\cite{lin2014coco}
(Table~\ref{tab:supp-impl}). The document U-Net is trained on RVL-CDIP~\cite{harley2015rvlcdip}
(Sec.~\ref{sec:supp-documents}). On Aachen, the solvers are adapted on the database images
(Sec.~\ref{sec:supp-aachen}).

\customparagraph{Testing.}
Unless stated otherwise, results are on the $200$ held-out COCO test images.
Table~\ref{tab:supp-testsets} lists the other test sets, which we process the same way. Localization is evaluated
on Aachen Day--Night v1.0~\cite{sattler2018benchmarking}, with its $824$ day and $98$ night queries
(Sec.~\ref{sec:supp-aachen}).

\begin{table}[!h]
  \centering
  \footnotesize
  \setlength{\tabcolsep}{3pt}
  \begin{tabularx}{\linewidth}{@{}lrXr@{}}
    \toprule
    \textbf{Set} & \textbf{Images} & \textbf{Which images} & \textbf{Keypoints} \\
    \midrule
    COCO & $200$ & held-out test images & $1617$ \\
    MegaDepth & $806$ & both MegaDepth-1500 test scenes & $1177$ \\
    DIV2K & $300$ & training images 0001--0300 & $1943$ \\
    ADE20K & $500$ & validation images 1--500 & $1162$ \\
    CelebA-HQ & $200$ & first $200$ training faces & $1528$ \\
    FUNSD & $50$ & all test forms & $1300$ \\
    \bottomrule
  \end{tabularx}
  \caption{\textbf{Test sets.} Keypoints: median ALIKED keypoints per image. Subsets are the first
    images in each dataset's order, not selected by content.}
  \label{tab:supp-testsets}
\end{table}

\section{Feature extractors}
\label{sec:supp-extractors}\suppfrozen{3}

We study five learned extractors with upright descriptors: ALIKED~\cite{zhao2023aliked}, XFeat~\cite{potje2024xfeat},
SuperPoint~\cite{detone2018superpoint}, DISK~\cite{tyszkiewicz2020disk} and LoMa~\cite{nordstrom2026loma}. Their image
regions are aligned with the image axes, and they omit explicit per-keypoint scale normalization, so neighboring
descriptors describe overlapping regions in a common frame. Their descriptors have $128$ (ALIKED, DISK), $64$
(XFeat) or $256$ (SuperPoint, LoMa) dimensions. ALIKED is our main extractor, and Table~\ref{tab:main} evaluates all
five. SIFT~\cite{lowe2004sift} instead rotates and rescales each region to its keypoint's orientation and scale, and
is treated on its own (Sec.~\ref{sec:exp-sift} and Sec.~\ref{sec:supp-sift}). Sec.~\ref{sec:supp-impl} gives the
extraction settings, and Sec.~\ref{sec:supp-mechanism} measures how far an ALIKED descriptor reaches.

\section{Mechanism}
\label{sec:supp-mechanism}\suppfrozen{4}

\customparagraph{Measurements of Fig.~\ref{fig:mechanism-method}.}
For the influence map in the middle, the probe photograph stays fixed on a white background, and we paste a
small textured block at each point of a grid around it. For every block position we recompute the probe's
descriptors at fixed keypoints and record how much they change ($1-\cos$). The map averages three keypoints of
the probe, at its left edge, center and right edge, over five probe photographs. The two circles mark the
radii $r_{50}$ and $r_{90}$ that enclose half and $90\%$ of a keypoint's descriptor change ($31$ and $84$\,px,
medians over keypoints). For the curve on the right, each trial takes the median change over the probe's keypoints and the
eight neighbor directions, and the line and band show the median and interquartile range over ten trials. The
horizontal axis is the grid spacing, so diagonal neighbors are farther away.

\customparagraph{Upright and oriented SIFT at the same keypoints.}
To isolate the orientation convention, we compute SIFT~\cite{lowe2004sift} with one large, fixed support at
the ALIKED keypoints of each image and train the solver twice, once on upright and once on oriented
descriptors, with nothing else changed. The solver places the upright descriptors to $12$\,px but fails on the
oriented ones ($141$\,px). The frame alone removes most of what the solver recovers. Native SIFT, whose
support grows with the keypoint's scale, is studied in Sec.~\ref{sec:supp-sift}.

\begin{figure}[t]
  \centering
  \includegraphics[width=\linewidth, trim=0 42bp 168bp 0, clip]{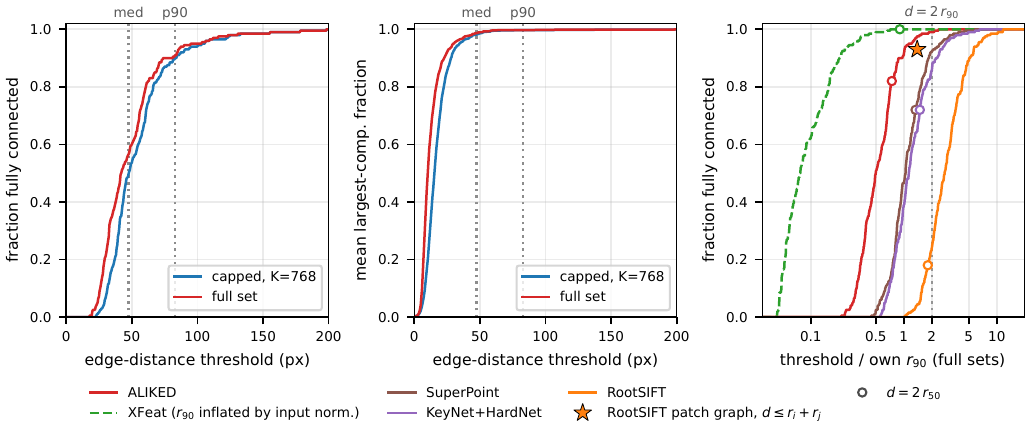}
  \caption{\textbf{Keypoints of real images connect at small distances.} ALIKED keypoints of the $200$ test images,
    linked when closer than the threshold. Left: fraction of images whose keypoints form one component.
    Right: mean fraction of keypoints in the largest component. Blue: $768$ sampled keypoints, as the solver
    sees them; red: all detected keypoints. Dotted: median and $90$th percentile of the threshold at which a
    sampled set becomes connected ($47.6$ and $83.0$\,px).}
  \label{fig:supp-connectivity}
\end{figure}

\customparagraph{Connectivity.}
\label{sec:supp-connectivity}
Two distant regions of an image can be linked through the keypoints between them (Sec.~\ref{sec:mechanism}).
Fig.~\ref{fig:supp-connectivity} measures how soon the ALIKED keypoints of a test image form one connected
graph when every pair closer than a threshold is linked. We use all detected keypoints, or a random sample of
them as the solver sees them. A typical sampled set is connected at about $50$\,px, and the largest component
holds most keypoints much earlier. The graph uses positions only. It shows that chains between distant regions
exist in real images, and Fig.~\ref{fig:mechanism-method} shows that neighboring descriptors share information
along them.

\section{Reference reconstructions}
\label{sec:supp-reference}\suppfrozen{5}

Image scores need a reference, and we use two. Tables~\ref{tab:image} and~\ref{tab:supp-ood-image} compare the
reconstructions from recovered and from random positions with the \emph{reference reconstruction}, made from the ground-truth (GT)
positions with the same descriptors and the same inverter (Sec.~\ref{sec:metrics}). The two reconstructions differ
only in the positions, so these scores show what hiding the positions protects, separately from what the inverter
can reconstruct at all. The italic rows of both tables give that second part: they compare the reference with the
original image, so they depend only on the inverter.

All other image scores compare each reconstruction with the original image and give the reconstructions from GT and
from recovered positions side by side: Table~\ref{tab:loc} and Tables~\ref{tab:supp-spaced},
\ref{tab:supp-aachen-query}, \ref{tab:supp-benchmark} and~\ref{tab:supp-precision}. These experiments change how
many descriptors are released or how precise their positions are, so a reconstruction from GT positions would not be
a fixed reference. With few descriptors it loses most objects itself, and two poor reconstructions can agree on the
same generic objects (Sec.~\ref{sec:supp-aachen}). Faces are identified among the original images
(Table~\ref{tab:facereid}).

\section{Metrics}
\label{sec:supp-metrics}\suppfrozen{6}

\customparagraph{Layouts.}
Each metric is computed per image, and we report its median over the test images (Sec.~\ref{sec:metrics}). The
residual is the median distance between recovered and GT positions after aligning the layouts, by translation
for upright descriptors and by rotation and translation for oriented ones. The spacing, the median distance from
a GT keypoint to its nearest neighbor, puts the residual in context. dcorr is the Pearson correlation between
recovered and GT distances over all keypoint pairs. kNN is the fraction of each keypoint's five nearest GT
neighbors that are also among its five nearest recovered neighbors, averaged over keypoints. Both are $1$ for a
perfect layout and close to $0$ for random positions. The MDS and Isomap baselines are aligned by a similarity
transform instead (Sec.~\ref{sec:supp-impl}).

\customparagraph{Images.}
PSNR and SSIM are computed on color images and LPIPS~\cite{zhang2018lpips} on grayscale images. Sec.~\ref{sec:supp-reference}
says which tables compare with the reconstruction from GT positions and which with the original image.

\customparagraph{Object recovery.}
OR follows the benchmark of~\cite{pietrantoni2026vulnerability}. Qwen2-VL~\cite{wang2024qwen2vl} lists up to
ten objects in each image, leaving out walls, floors, sky, doors and trees. OR is the fraction of the
objects listed for the reference image, the reconstruction from GT positions or the original image, that are also
listed for the reconstruction, matched one to one by the similarity of their
MiniLM sentence embeddings. Two more prompts of the benchmark give texture recall (TR) and scene similarity
(SR), reported in Table~\ref{tab:supp-benchmark}. Decoding is greedy, so the scores are deterministic.

\customparagraph{Faces and text.}
On faces, ArcFace~\cite{deng2019arcface} identifies each reconstruction among the original images
(Table~\ref{tab:facereid}). On documents, word recall is the share of a form's words that appear in a model's
transcription of the reconstruction (Sec.~\ref{sec:supp-documents}).

\section{Recovery and inversion across domains}
\label{sec:supp-oodres}\suppfrozen{7}

Sec.~\ref{sec:exp-recovery} recovers keypoint positions on COCO and on five domains not seen in training
(Table~\ref{tab:main}), and Sec.~\ref{sec:exp-inversion} reconstructs images from them (Table~\ref{tab:image},
Fig.~\ref{fig:inversion}). This section shows more layouts and reconstructions for the four image domains. Forms are
in Sec.~\ref{sec:supp-documents}.

\customparagraph{What the image scores compare.}
In Tables~\ref{tab:image} and~\ref{tab:supp-ood-image}, each test image is reconstructed from the same descriptors
with the same inverter, once at the GT positions and once at the recovered positions. PSNR, SSIM, LPIPS and object
recovery (OR) compare the reconstruction from recovered positions, or from random positions, with the
reconstruction from GT positions, not with the original image. An OR of $1$ thus means that hiding the positions
removed no object the inverter could otherwise show. The italic rows (GT vs.\ original) compare the reconstruction
from GT positions with the original image and measure the inverter itself (Sec.~\ref{sec:supp-reference}).

\begin{figure*}[t]
  \centering
  \begin{minipage}[t]{0.48\linewidth}%
    \centering
    \includegraphics[width=\linewidth]{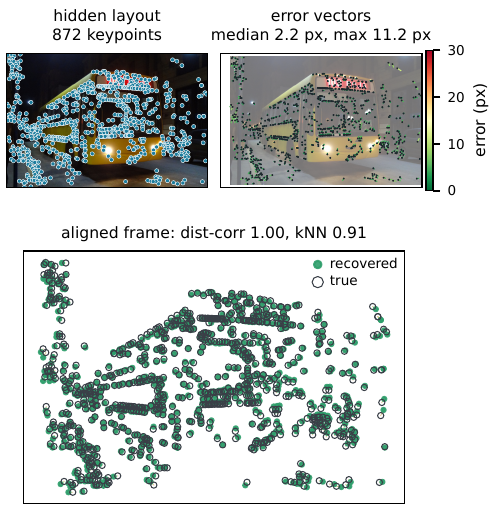}\\[3pt]
    {\footnotesize (a) best: $2.2$\,px}%
  \end{minipage}\hfill
  \begin{minipage}[t]{0.48\linewidth}%
    \centering
    \includegraphics[width=\linewidth]{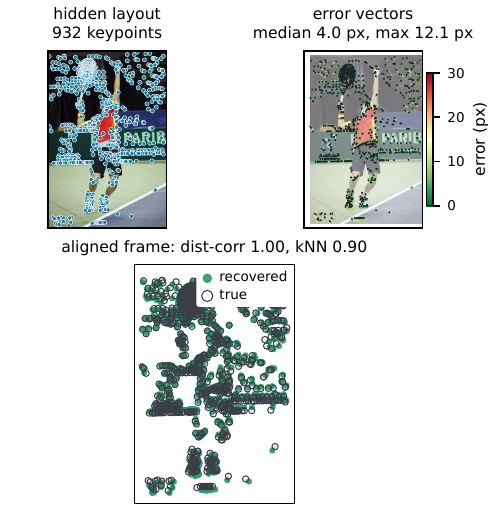}\\[3pt]
    {\footnotesize (b) median: $4.0$\,px}%
  \end{minipage}

  \vspace{6pt}
  \begin{minipage}[t]{0.48\linewidth}%
    \centering
    \includegraphics[width=\linewidth]{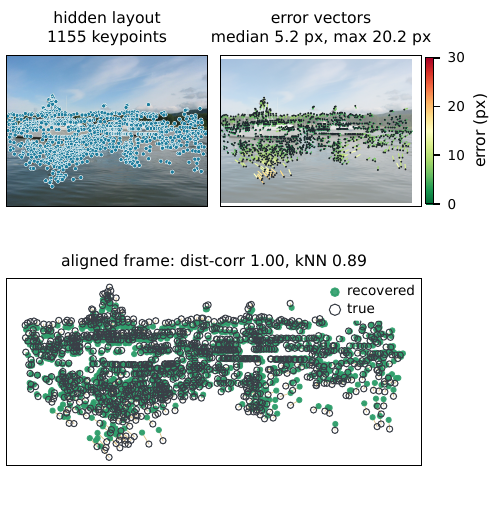}\\[3pt]
    {\footnotesize (c) 75th pct.: $5.2$\,px}%
  \end{minipage}\hfill
  \begin{minipage}[t]{0.48\linewidth}%
    \centering
    \includegraphics[width=\linewidth]{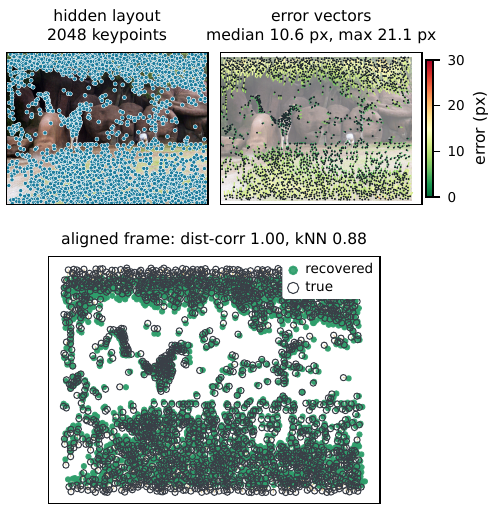}\\[3pt]
    {\footnotesize (d) worst of $120$ scanned: $10.6$\,px}%
  \end{minipage}
  \caption{\textbf{Recovered layouts across the error distribution.} GT keypoints in gray, recovered in green, after
    aligning the translation: the best, median, $75$th-percentile and worst of $120$ test images, with
    their median residuals.}
  \label{fig:supp-cases}
\end{figure*}

\customparagraph{Single images.}
Fig.~\ref{fig:supp-cases} shows recovered layouts from the best to the worst of $120$ test images. Even in the
worst case, the arrangement of the scene holds.

\begin{figure}[t]
  \centering
  \includegraphics[width=\linewidth]{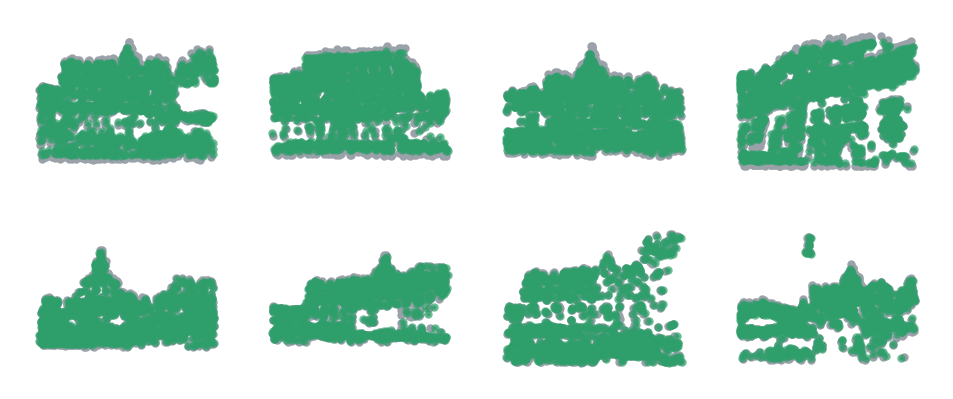}\\[3pt]
  \includegraphics[width=\linewidth]{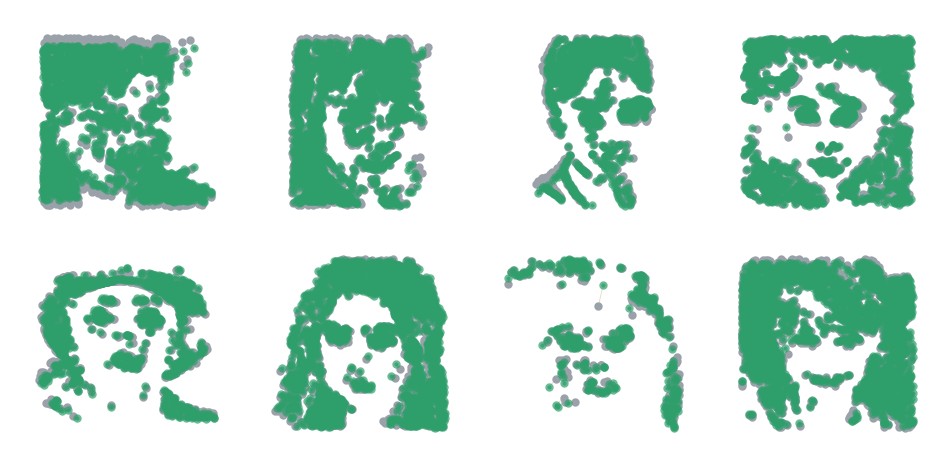}
  \caption{\textbf{Recovered layouts on landmarks and faces.} Recovered (green) over GT (gray) keypoints, with the
    solver trained on COCO. Landmarks (MegaDepth, top) are recovered as precisely as COCO; faces (CelebA-HQ,
    bottom) less precisely, but their arrangement holds.}
  \label{fig:supp-ood-domains}
\end{figure}

\begin{figure}[t]
  \centering
  \includegraphics[width=\linewidth]{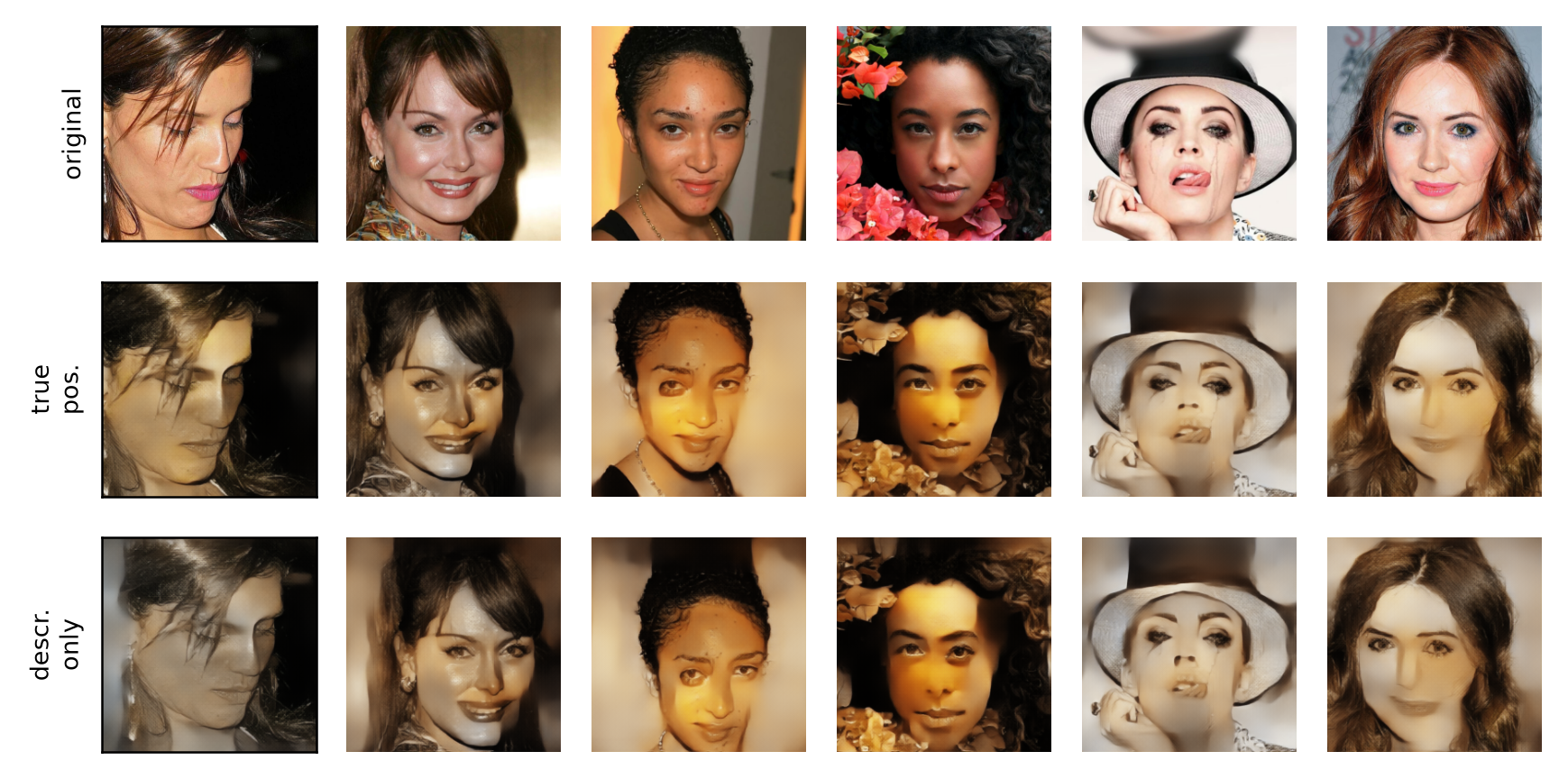}
  \caption{\textbf{Faces from descriptors alone.} CelebA-HQ, with the solver and the U-Net trained on COCO: original
    (top), reconstruction from GT positions (middle) and from recovered positions (bottom).}
  \label{fig:ood-faces}
\end{figure}

\begin{figure*}[t]
  \centering
  \includegraphics[width=0.85\linewidth]{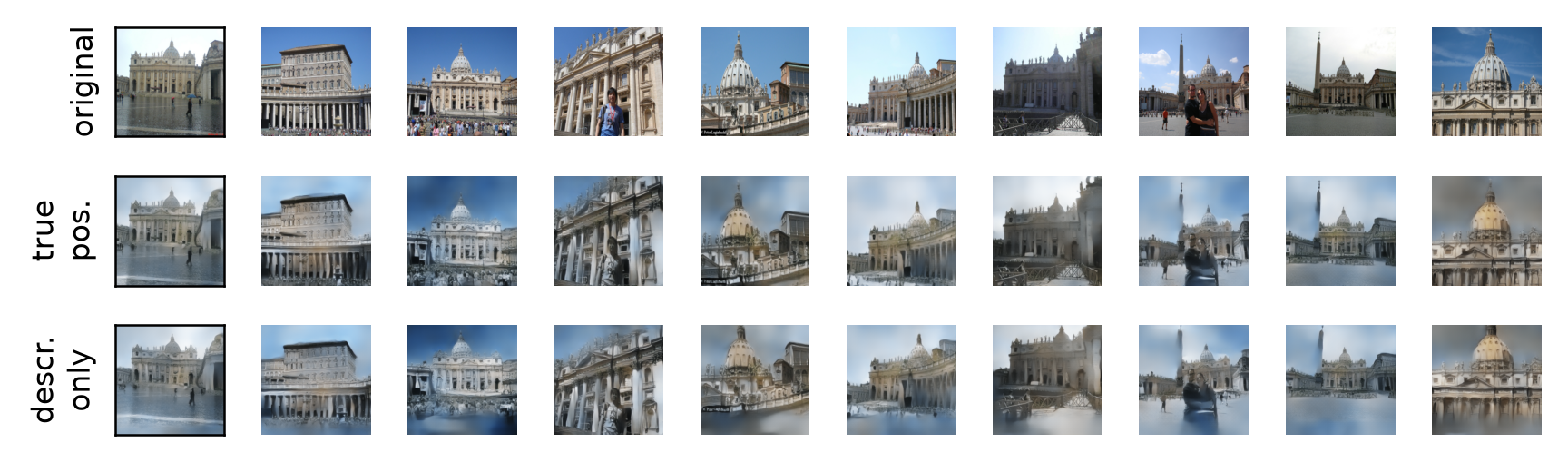}
  \caption{\textbf{Landmarks from descriptors alone.} MegaDepth, with the solver and the U-Net trained on COCO:
    original (top), reconstruction from GT positions (middle) and from recovered positions (bottom).}
  \label{fig:supp-ood-recon}
\end{figure*}

\customparagraph{Other domains.}
Fig.~\ref{fig:supp-ood-domains} overlays recovered and GT layouts for landmarks and faces, two of the
domains of Table~\ref{tab:main}. Table~\ref{tab:supp-ood-image} repeats Table~\ref{tab:image} on the first
$100$ images of each domain, with the solver and the U-Net trained on COCO only. As on COCO, the reconstruction
from recovered positions stays close to the one from GT positions, and random positions recover few objects. Fig.~\ref{fig:inversion} shows, for each domain, an image within two ranks of
the median residual, and Figs.~\ref{fig:ood-faces} and~\ref{fig:supp-ood-recon} show more.

\begin{table}[t]
  \centering
  \footnotesize
  \setlength{\tabcolsep}{3pt}
  \begin{tabular*}{\linewidth}{@{\extracolsep{\fill}}llcccc@{}}
    \toprule
    \textbf{Domain} & \textbf{Positions} & \textbf{PSNR} $\uparrow$ & \textbf{SSIM} $\uparrow$ & \textbf{LPIPS} $\downarrow$ & \textbf{OR} $\uparrow$ \\
    \midrule
    MegaDepth & recovered & 21.2 & 0.66 & 0.11 & 0.64 \\
              & random    & 12.7 & 0.19 & 0.60 & 0.20 \\
              & \textit{GT vs.\ original} & \textit{17.5} & \textit{0.75} & \textit{0.19} & \textit{0.58} \\
    \midrule
    DIV2K     & recovered & 20.5 & 0.60 & 0.11 & 0.62 \\
              & random    & 12.1 & 0.19 & 0.59 & 0.08 \\
              & \textit{GT vs.\ original} & \textit{17.4} & \textit{0.67} & \textit{0.24} & \textit{0.51} \\
    \midrule
    ADE20K    & recovered & 21.8 & 0.70 & 0.11 & 0.69 \\
              & random    & 12.0 & 0.24 & 0.60 & 0.09 \\
              & \textit{GT vs.\ original} & \textit{17.1} & \textit{0.70} & \textit{0.23} & \textit{0.55} \\
    \midrule
    CelebA-HQ & recovered & 22.1 & 0.73 & 0.09 & 0.65 \\
              & random    & 10.8 & 0.24 & 0.62 & 0.08 \\
              & \textit{GT vs.\ original} & \textit{16.6} & \textit{0.69} & \textit{0.22} & \textit{0.52} \\
    \bottomrule
  \end{tabular*}
  \caption{\textbf{Table~\ref{tab:image} on the other domains.} Our U-Net and the ALIKED solver, both trained on COCO
    only, on the first $100$ images of each domain. As in Table~\ref{tab:image}, recovered and random
    positions are scored against the reconstruction from GT positions, and the italic rows score that
    reference against the original image.}
  \label{tab:supp-ood-image}
\end{table}

\customparagraph{Diffusion inverter.}
For the diffusion rows of Table~\ref{tab:image}, we fine-tune the conditional diffusion inverter
of~\cite{pietrantoni2026vulnerability}, released for SIFT, SuperPoint and XFeat, on upright ALIKED for
$50$k steps. %

\customparagraph{Face identification.}
We run ArcFace~\cite{deng2019arcface} (SCRFD detector, w600k\_r50 embedding) on every CelebA-HQ reconstruction
and identify it among the $200$ originals (Table~\ref{tab:facereid}). From recovered positions, almost every
reconstruction contains a detected face, and almost all of these are identified correctly, close to the result
from GT positions. From random positions, no face is detected. Some people appear more than once in CelebA-HQ,
which can only lower these scores.

\begin{table}[t]
  \centering
  \footnotesize
  \setlength{\tabcolsep}{2pt}
  \begin{tabular*}{\linewidth}{@{\extracolsep{\fill}}lccccc@{}}
    \toprule
    \textbf{Positions} & \textbf{Det.} $\uparrow$ & \textbf{Gen.\ cos} $\uparrow$ & \textbf{AUC} $\uparrow$ & \textbf{Rank-1} $\uparrow$ & \textbf{TPR@$\mathbf{10^{-3}}$} $\uparrow$ \\
    \midrule
    GT        & 1.00 & 0.55 & 1.000 & 1.00 & 0.995 \\
    Recovered & 0.99 & 0.46 & 0.999 & 0.98 & 0.93 \\
    Random    & 0.00 & --  & --    & --   & --   \\
    \bottomrule
  \end{tabular*}
  \caption{\textbf{Face identification on CelebA-HQ.} ArcFace~\cite{deng2019arcface} on $200$ reconstructions,
    identified among the $200$ originals. Det.: fraction with a detected face; gen.\ cos: mean cosine
    similarity to the original (to other people: $0.01$); TPR at a false-accept rate of $10^{-3}$.}
  \label{tab:facereid}
\end{table}

\section{SIFT with released scale and orientation}
\label{sec:supp-sift}\suppfrozen{8}

\customparagraph{Release.}
Sec.~\ref{sec:exp-sift} attacks SIFT records that keep each keypoint's scale $\sigma$ and orientation
$\theta$, as OpenCV~\cite{opencv_library} keypoint files and COLMAP~\cite{schoenberger2016mvs,schoenberger2016sfm} feature files store them. Each descriptor is computed in the
frame rotated by $\theta$, and the attacker gets everything but the positions. Keypoints are OpenCV's
difference-of-Gaussians detections with a low contrast threshold, on images resized to $512$\,px on the longer
side, capped at the number of ALIKED keypoints of the same image. The solver is the one of
Sec.~\ref{sec:supp-impl}, trained on COCO images disjoint from the test images.

\customparagraph{Undoing the orientation.}
The released angle tells the attacker in which frame each descriptor was computed. A small MLP maps the
oriented descriptor and $(\cos\theta, \sin\theta)$ to the upright descriptor at the same keypoint. It is
trained with a cosine loss on pairs computed from the attacker's own images. A fixed rearrangement of the
descriptor would not work, because SIFT interpolates between its spatial and orientation bins. On the test
keypoints, the MLP's output is close to the true upright descriptor (median cosine $0.99$, against $0.62$ for
the released one).

\customparagraph{Scale.}
We append $\log\sigma$ to each descriptor as one more input dimension, standardized and scaled to the
spread of a RootSIFT entry.

\begin{figure}[t]
  \centering
  \includegraphics[width=\linewidth]{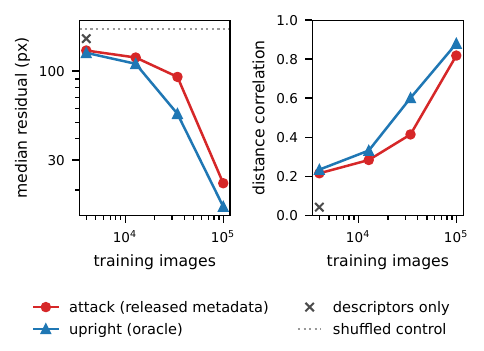}
  \caption{\textbf{The SIFT attack improves with training data.} Median residual (left) and dcorr (right) against the
    number of training images. Red: the attack, with the released descriptors de-rotated by the MLP and the
    released scale. Blue, the oracle: descriptors computed upright from the image, which the attacker does not have, with the
    released scale. It is the best case, a lower bound on the residual (left) and an upper bound on dcorr (right).
    Crosses: the released descriptors without scale; dotted: the solver given the descriptors in random order.}
  \label{fig:released-metadata-scaling}
\end{figure}

\begin{figure*}[t]
  \centering
  \includegraphics[width=0.55\textwidth, trim=0 0 127bp 0, clip]{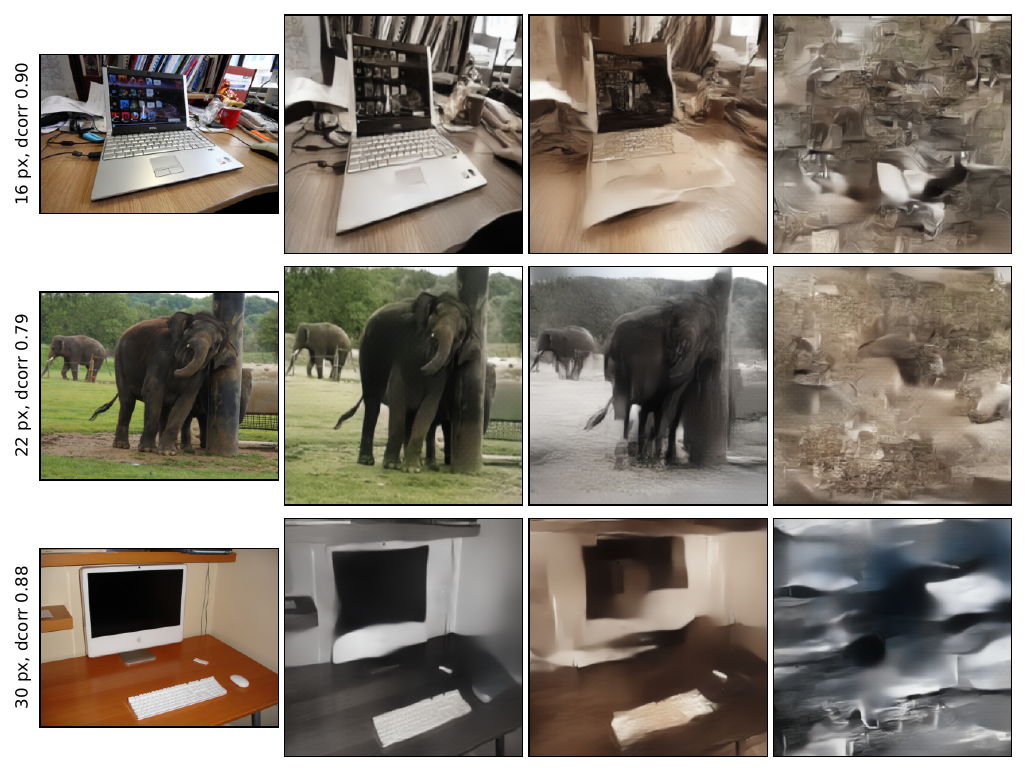}
  \caption{\textbf{Reconstructions from the SIFT attack.} Test images at the $25$th, $50$th and $75$th percentile of
    the residual ($100$k training images); row labels give residual and dcorr. Columns: original; U-Net at GT
    positions with upright descriptors; U-Net at the recovered positions with the de-rotated descriptors.}
  \label{fig:released-metadata-render}
\end{figure*}

\customparagraph{Results.}
Fig.~\ref{fig:released-metadata-scaling} shows that the attack improves with more training data. With $100$k
training images it reaches the $22$\,px of Sec.~\ref{sec:exp-sift}, and the curves have not yet flattened.
The oracle, with descriptors computed upright from the image, which the attacker does not have, does somewhat
better, so undoing the orientation costs a few pixels. Without the scale, the solver recovers nothing, from the released and from
the upright descriptors alike. Fig.~\ref{fig:released-metadata-render} shows reconstructions by a U-Net
trained with our recipe on upright native-scale SIFT at GT positions.

\section{Defenses on COCO}
\label{sec:supp-defenses}\suppfrozen{9}

Sec.~\ref{sec:defenses} asks whether a client can protect the layout by changing the descriptors it sends, for
example by sending fewer of them or quantizing them, while they stay useful for localization. This section gives
the details of those experiments and tests two more modifications: releasing only keypoints far apart
(Sec.~\ref{sec:supp-spaced}) and hiding the descriptors among unrelated ones (Sec.~\ref{sec:supp-outliers}).

\customparagraph{Adapted and unadapted solvers.}
A modification protects the layout only if it holds against an attacker who knows it. Such an attacker can apply
the same modification to its own public training images and fine-tune the solver on them. We call this solver
\emph{adapted}, following the practice of adaptive evaluation~\cite{carlini2019evaluating,tramer2020adaptive}. The
\emph{unadapted} solver is fine-tuned for the same number of epochs on unmodified descriptors and then applied to
modified ones. It shows what the modification does to an attacker who ignores it. On COCO, both solvers start from the ALIKED solver
of Table~\ref{tab:main} and train for $10$ more epochs at a lower learning rate. The image panels of Fig.~\ref{fig:defense-inversion} show one test
image near the median, inverted by the U-Net trained on unmodified descriptors. On Aachen
(Sec.~\ref{sec:supp-aachen}), the unadapted and the adapted query solvers start from the LoMa solver of
Table~\ref{tab:main} and train for $35$ more epochs on the database images, released unmodified or modified like the
queries. They are tested on the queries.

\subsection{Farthest-first spatial selection}
\label{sec:supp-spaced}

If descriptors leak through the image regions they share, releasing only keypoints far from each other might
help. How far apart is far depends on the extractor. We use ALIKED (Sec.~\ref{sec:supp-extractors}), for which half
of the change in a descriptor comes from content within $r_{50}=31$\,px of its keypoint and $90\%$ from content
within $r_{90}=84$\,px (Sec.~\ref{sec:supp-mechanism}). Even keypoints $128$\,px apart, the largest spacing we test,
therefore still describe overlapping regions. We select keypoints farthest-first from a random start, each at least $d$ from those already kept, and
compare with a random subset of the same size (Table~\ref{tab:supp-spaced}). Each release gets its own solver,
adapted as above, and the U-Net inverts the first $100$ test images.

\begin{figure*}[t]
  \centering
  \includegraphics[width=\linewidth]{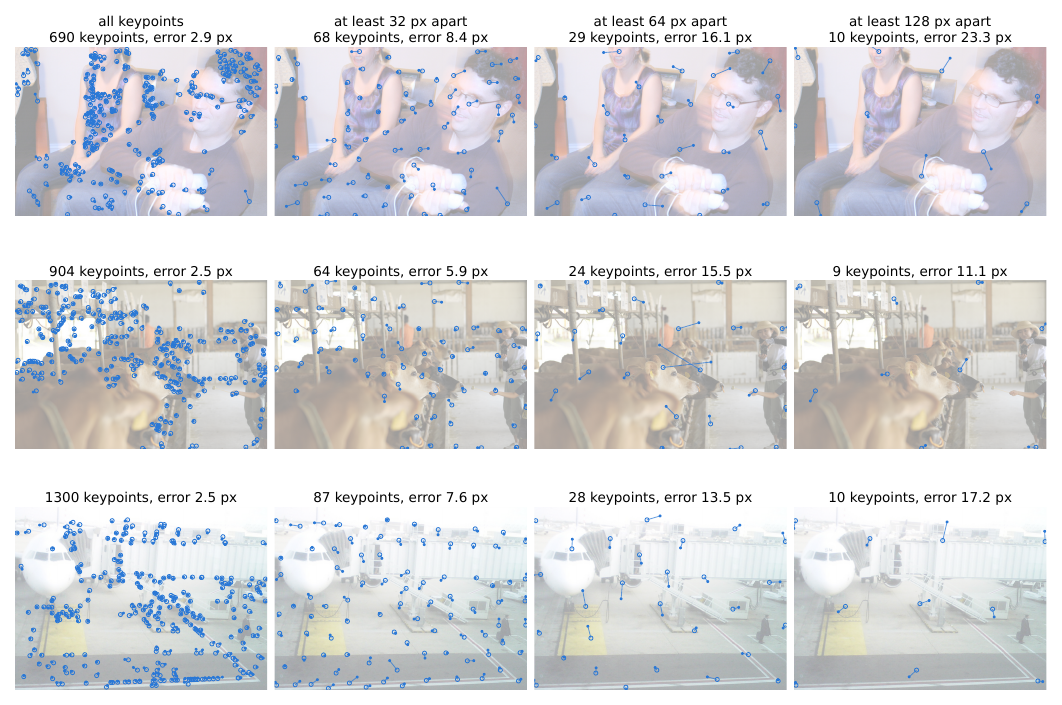}
  \caption{\textbf{Spaced releases on three test images.} Each column is attacked by the solver adapted to that
    release. Circles: GT keypoints; dots: recovered keypoints; lines: errors. Titles give the number of
    keypoints and the median residual; at most $300$ keypoints are drawn.}
  \label{fig:supp-spaced}
\end{figure*}

Spreading the keypoints does not protect the layout. At every distance, the spaced release is recovered at
least as well as the random one, and even the sparsest releases are placed near their GT positions
(Fig.~\ref{fig:supp-spaced}). Sparse releases lose content instead. With fewer than about $100$ descriptors,
the U-Net no longer draws recognizable objects, even at the GT positions. %

\begin{table}[t]
  \centering
  \footnotesize
  \setlength{\tabcolsep}{3.2pt}
  \begin{tabular*}{\linewidth}{@{\extracolsep{\fill}}lccccccc@{}}
    \toprule
    & & \multicolumn{2}{c}{\textbf{Spaced}} & \multicolumn{2}{c}{\textbf{Random}} & \multicolumn{2}{c}{\textbf{OR, spaced} $\uparrow$} \\
    \cmidrule(lr){3-4}\cmidrule(lr){5-6}\cmidrule(lr){7-8}
    \textbf{$\boldsymbol{d}$ (px)} & \textbf{Kept} & \textbf{px} $\downarrow$ & \textbf{dcorr} $\uparrow$ & \textbf{px} $\downarrow$ & \textbf{dcorr} $\uparrow$ & \textbf{GT} & \textbf{Rec.} \\
    \midrule
    -- & $1617$ & $3.5$ & $0.999$ & -- & -- & $0.56$ & $0.52$ \\
    $8$ & $601$ & $3.4$ & $0.999$ & $3.8$ & $0.999$ & $0.49$ & $0.46$ \\
    $16$ & $240$ & $4.3$ & $0.999$ & $5.7$ & $0.998$ & $0.28$ & $0.26$ \\
    $32$ & $84$ & $7.6$ & $0.996$ & $10.1$ & $0.992$ & $0.09$ & $0.08$ \\
    $64$ & $29$ & $15.2$ & $0.983$ & $19.6$ & $0.966$ & $0.03$ & $0.03$ \\
    $96$ & $16$ & $20.3$ & $0.970$ & $28.1$ & $0.944$ & $0.04$ & $0.03$ \\
    $128$ & $10$ & $21.9$ & $0.973$ & $32.2$ & $0.926$ & $0.03$ & $0.05$ \\
    \bottomrule
  \end{tabular*}
  \caption{\textbf{Spaced releases of ALIKED descriptors on COCO.} $d$: minimum distance between released keypoints ($512$\,px frame);
    kept: median number of keypoints. Residual and dcorr of the solver adapted to each release, for the
    spaced release and for a random one of the same size. OR: U-Net at GT and at recovered positions
    (random positions: $0.05$). The first row releases all keypoints. Images left with fewer than $8$
    keypoints are skipped ($23$ at $d{=}128$\,px).}
  \label{tab:supp-spaced}
\end{table}

\subsection{Distractor descriptors}
\label{sec:supp-outliers}

A client could also hide its descriptors among those of other images. We add descriptors of other test images,
either drawn from many images or taken from whole images, which form coherent scenes, up to ten times as many
as the image's own (Table~\ref{tab:supp-outliers}). The ALIKED solver runs on all descriptors together,
without retraining, and we score the image's own keypoints. The solver also predicts a variance per keypoint,
so an attacker who knows how many descriptors were added can drop that many of the most uncertain ones and run
the solver again.

\begin{table}[t]
  \centering
  \footnotesize
  \setlength{\tabcolsep}{3.2pt}
  \begin{tabular*}{\linewidth}{@{\extracolsep{\fill}}lcccccccc@{}}
    \toprule
    & \multicolumn{4}{c}{\textbf{From many images}} & \multicolumn{4}{c}{\textbf{From whole images}} \\
    \cmidrule(lr){2-5}\cmidrule(lr){6-9}
    \textbf{Added} & \textbf{px} $\downarrow$ & \textbf{AUC} $\uparrow$ & \textbf{Filt.} $\downarrow$ & \textbf{Kept} & \textbf{px} $\downarrow$ & \textbf{AUC} $\uparrow$ & \textbf{Filt.} $\downarrow$ & \textbf{Kept} \\
    \midrule
    -- & $4.0$ & -- & -- & -- & $4.0$ & -- & -- & -- \\
    $10\%$ & $5.1$ & $0.97$ & $3.9$ & $0.98$ & $5.2$ & $0.94$ & $4.4$ & $0.96$ \\
    $25\%$ & $16.0$ & $0.97$ & $3.8$ & $0.96$ & $7.6$ & $0.77$ & $7.4$ & $0.87$ \\
    $50\%$ & $34.7$ & $0.96$ & $4.5$ & $0.93$ & $10.7$ & $0.63$ & $10.1$ & $0.73$ \\
    $100\%$ & $43.7$ & $0.96$ & $6.1$ & $0.88$ & $11.1$ & $0.57$ & $10.9$ & $0.59$ \\
    $200\%$ & $47.8$ & $0.95$ & $8.1$ & $0.84$ & $15.8$ & $0.54$ & $13.8$ & $0.41$ \\
    $300\%$ & $50.6$ & $0.95$ & $9.6$ & $0.81$ & $20.6$ & $0.51$ & $15.7$ & $0.30$ \\
    $500\%$ & $50.9$ & $0.94$ & $13.4$ & $0.77$ & $24.6$ & $0.52$ & $16.6$ & $0.21$ \\
    $1000\%$ & $50.6$ & $0.92$ & $26.3$ & $0.65$ & $29.8$ & $0.53$ & $16.9$ & $0.12$ \\
    \bottomrule
  \end{tabular*}
  \caption{\textbf{Distractor descriptors added to a release.} Added: descriptors of other test images, as a share of
    the image's own count. px: median residual of the image's own keypoints for the ALIKED solver on all
    descriptors together. AUC: how well the solver's variance separates added from own descriptors. Filt.\ and kept: residual
    after dropping as many of the most uncertain descriptors as were added and running the solver again, and
    the share of own keypoints left. Images left with fewer than $8$ keypoints count as failures ($3$, $6$
    and $23$ of $200$ at $300$, $500$ and $1000\%$ from whole images).}
  \label{tab:supp-outliers}
\end{table}

\customparagraph{Descriptors from many images.}
These distractors disturb the solver, but its variance separates them from the image's own descriptors. After
filtering, the error of the image's own keypoints is close to that of the clean release, as long as no more
descriptors are added than the image has.

\customparagraph{Whole scenes.}
Whole scenes cannot be separated this way, but they disturb the solver less. One to three added scenes raise
the error only to $11$ to $21$\,px, and even ten added scenes leave it around $30$\,px.

\customparagraph{Comparison with LDP-Feat.}
LDP-Feat~\cite{pittaluga2023ldpfeat} also hides each descriptor among unrelated ones, but unlike our distractors
it changes the descriptor itself. It replaces the descriptor by its nearest word in a public dictionary, built by
$k$-means clustering of descriptors from other images, and sends $m$ words per keypoint. The keypoint's own word
is among them only with a small probability, and otherwise all $m$ words are random. We apply LDP-Feat to the
COCO test images with its default privacy level, $\varepsilon = 10$, and a dictionary of $256$k words built from
the ALIKED descriptors of other COCO images. Only $8$ to $26\%$ of the keypoints then keep their own word
($m = 1$ to $4$).

The two defenses behave differently. Distractors leave the image's own descriptors unchanged, so the ALIKED solver
still recovers their positions (Table~\ref{tab:supp-outliers}). Under LDP-Feat, the solver does not recover
the positions, not even for the keypoints that kept their own word, and its variance does not single out the true
words (Table~\ref{tab:supp-ldp}). The solver also fails when every keypoint sends only its own word, so the random words
are not what stops it. An attacker who knows the defense does not get much further. We fine-tune the
solver as in Sec.~\ref{sec:supp-defenses}, once on own words and once on LDP-Feat's word sets. On own words it then
recovers only a coarse layout, and on the word sets the keypoints that kept their own word stay far from their
positions with either adapted solver. Replacing the descriptors by dictionary words protects the layout, also against
an adapted solver, while hiding unchanged descriptors among others does not. The random words, however, cost
localization. On Aachen Day--Night with SIFT, LDP-Feat with $m = 2$ localizes $42\%$ of the day queries, against
$77\%$ when each descriptor is only replaced by its own word~\cite{pittaluga2023ldpfeat}.

\begin{table}[t]
  \centering
  \footnotesize
  \setlength{\tabcolsep}{3pt}
  \begin{tabular*}{\linewidth}{@{\extracolsep{\fill}}lccc@{}}
    \toprule
    & & \multicolumn{2}{c}{\textbf{Adapted to}} \\
    \cmidrule(l){3-4}
    \textbf{Release} & \textbf{Unadapted} & \textbf{own words} & \textbf{LDP-Feat, $\boldsymbol{m{=}2}$} \\
    \midrule
    Own words & $164$ & $97$ & $157$ \\
    LDP-Feat, $m{=}1$ & $198$ & $147$ & $173$ \\
    LDP-Feat, $m{=}2$ & $190$ & $137$ & $166$ \\
    LDP-Feat, $m{=}4$ & $175$ & $133$ & $167$ \\
    \bottomrule
  \end{tabular*}
  \caption{\textbf{LDP-Feat hides the layout, also from adapted solvers.} Median distance (px) between the keypoints
    that sent their own word and their GT positions, after aligning the translation, on the $200$ COCO test images
    ($\varepsilon = 10$, $256$k words). The adapted solvers are fine-tuned as in Sec.~\ref{sec:supp-defenses}, on
    $13{,}800$ training images. Without LDP-Feat the solver reaches $3.9$\,px; placing every keypoint at the image
    center gives $182$\,px and random positions $231$\,px.}
  \label{tab:supp-ldp}
\end{table}

\section{Defenses on Aachen Day--Night}
\label{sec:supp-aachen}\suppfrozen{10}

\begin{table*}[t]
  \centering
  \footnotesize
  \setlength{\tabcolsep}{3pt}
  \begin{tabular*}{\textwidth}{@{\extracolsep{\fill}}lcccccc@{}}
    \toprule
    & & \textbf{Loc.\ (\%)} $\uparrow$ & \multicolumn{2}{c}{\textbf{Error (px)} $\downarrow$ \textbf{/ dcorr} $\uparrow$} & \multicolumn{2}{c}{\textbf{OR, GT / attack} $\uparrow$} \\
    \cmidrule(lr){3-3}\cmidrule(lr){4-5}\cmidrule(lr){6-7}
    \textbf{Query release} & \textbf{B} & \textbf{day / night} & \textbf{unadapted} & \textbf{adapted} & \textbf{diffusion} & \textbf{U-Net} \\
    \midrule
    $5{,}000$ keypoints & $1024$ & $88.2$ / $90.8$ & $8.9$ / $0.998$ & $8.9$ / $0.998$ & $0.63$ / $0.64$ & $0.57$ / $0.56$ \\
    $1{,}024$ keypoints & $1024$ & $87.3$ / $82.7$ & $10.6$ / $0.997$ & $9.0$ / $0.998$ & $0.62$ / $0.58$ & $0.50$ / $0.47$ \\
    $512$ keypoints & $1024$ & $86.0$ / $79.6$ & $14.0$ / $0.995$ & $10.6$ / $0.997$ & $0.51$ / $0.51$ & $0.38$ / $0.37$ \\
    $256$ keypoints & $1024$ & $83.5$ / $76.5$ & $21.9$ / $0.989$ & $13.1$ / $0.994$ & $0.41$ / $0.41$ & $0.27$ / $0.25$ \\
    $128$ keypoints & $1024$ & $78.2$ / $73.5$ & $41.7$ / $0.966$ & $16.2$ / $0.990$ & $0.31$ / $0.29$ & $0.21$ / $0.18$ \\
    $64$ keypoints & $1024$ & $71.6$ / $56.1$ & $71.1$ / $0.909$ & $21.3$ / $0.983$ & $0.22$ / $0.19$ & $0.14$ / $0.10$ \\
    \midrule
    ${\geq}50$\,px ($119$) & $1024$ & $61.0$ / $43.9$ & $18.5$ / $0.991$ & $13.7$ / $0.996$ & $0.12$ / $0.14$ & $0.16$ / $0.17$ \\
    ${\geq}100$\,px ($35$) & $1024$ & $24.1$ / $9.2$ & $72.9$ / $0.831$ & $25.5$ / $0.981$ & $0.11$ / $0.08$ & $0.00$ / $0.01$ \\
    ${\geq}200$\,px ($11$) & $1024$ & $2.2$ / $0.0$ & $140.0$ / $0.550$ & $50.6$ / $0.917$ & $0.13$ / $0.12$ & $0.00$ / $0.01$ \\
    \midrule
    OPQ $16{\times}8$ & $16$ & $82.5$ / $79.6$ & $26.7$ / $0.936$ & $15.5$ / $0.985$ & $0.56$ / $0.53$ & $0.53$ / $0.42$ \\
    OPQ $8{\times}8$ & $8$ & $61.2$ / $37.8$ & $86.6$ / $0.642$ & $27.7$ / $0.893$ & $0.54$ / $0.48$ & $0.49$ / $0.34$ \\
    OPQ $4{\times}8$ & $4$ & $15.4$ / $8.2$ & $159.9$ / $0.390$ & $76.9$ / $0.660$ & $0.50$ / $0.39$ & $0.48$ / $0.17$ \\
    \bottomrule
  \end{tabular*}
  \caption{\textbf{Aachen query releases.} All settings of Table~\ref{tab:loc}, with the unadapted query solver next to
    the adapted one and with both inverters. B: bytes per descriptor (for 256 dimensional LoMa descriptors). Loc.: queries localized within
    $25$\,cm and $2^\circ$, matched without positions against a $4$-bit map. Error: median residual over the
    $922$ queries ($800$\,px frame). OR: object recovery on $100$ queries from GT positions and from the
    adapted solver's layout (random positions: $0.06$ with the diffusion inverter, $0.14$ with the U-Net).
    Top: the $N$ highest-scoring keypoints. Middle: keypoints at least $D$\,px apart ($n$: median count). Bottom: all keypoints through an OPQ codec.}
  \label{tab:supp-aachen-query}
\end{table*}

\begin{table*}[t]
  \centering
  \footnotesize
  \setlength{\tabcolsep}{3pt}
  \begin{tabular*}{\textwidth}{@{\extracolsep{\fill}}llcccccccccccc@{}}
    \toprule
    & & \multicolumn{6}{c}{\textbf{Our U-Net}} & \multicolumn{6}{c}{\textbf{Diffusion inverter}~\cite{pietrantoni2026vulnerability}} \\
    \cmidrule(lr){3-8} \cmidrule(lr){9-14}
    \textbf{Data / release} & \textbf{Positions} & \textbf{SSIM} $\uparrow$ & \textbf{LPIPS} $\downarrow$ & \textbf{FID} $\downarrow$ & \textbf{OR} $\uparrow$ & \textbf{TR} $\uparrow$ & \textbf{SR} $\uparrow$
      & \textbf{SSIM} $\uparrow$ & \textbf{LPIPS} $\downarrow$ & \textbf{FID} $\downarrow$ & \textbf{OR} $\uparrow$ & \textbf{TR} $\uparrow$ & \textbf{SR} $\uparrow$ \\
    \midrule
    COCO & GT & 0.69 & 0.22 & 179 & 0.54 & 0.21 & 0.78 & 0.49 & 0.28 & 120 & 0.68 & 0.59 & 0.80 \\
     & recovered & 0.50 & 0.25 & 194 & 0.54 & 0.20 & 0.77 & 0.42 & 0.31 & 124 & 0.65 & 0.57 & 0.81 \\
     & random & 0.16 & 0.62 & 342 & 0.06 & 0.00 & 0.42 & 0.15 & 0.66 & 364 & 0.05 & 0.01 & 0.38 \\
    \midrule
    $5{,}000$ keypoints & GT & 0.60 & 0.38 & 119 & 0.57 & 0.20 & 0.77 & 0.49 & 0.29 & 94 & 0.63 & 0.52 & 0.79 \\
     & recovered & 0.43 & 0.44 & 146 & 0.56 & 0.18 & 0.76 & 0.37 & 0.34 & 96 & 0.64 & 0.56 & 0.78 \\
     & random & 0.22 & 0.70 & 358 & 0.14 & 0.01 & 0.50 & 0.09 & 0.68 & 369 & 0.06 & 0.01 & 0.46 \\
    \noalign{\vskip 1.5pt}\cdashline{1-14}[1pt/1.5pt]\noalign{\vskip 1.5pt}
    $1{,}024$ keypoints & GT & 0.52 & 0.46 & 170 & 0.50 & 0.06 & 0.73 & 0.42 & 0.34 & 93 & 0.62 & 0.51 & 0.77 \\
     & recovered & 0.40 & 0.50 & 198 & 0.47 & 0.03 & 0.72 & 0.34 & 0.39 & 95 & 0.58 & 0.52 & 0.77 \\
    \noalign{\vskip 1.5pt}\cdashline{1-14}[1pt/1.5pt]\noalign{\vskip 1.5pt}
    $512$ keypoints & GT & 0.45 & 0.55 & 267 & 0.38 & 0.04 & 0.68 & 0.37 & 0.43 & 107 & 0.51 & 0.47 & 0.74 \\
     & recovered & 0.38 & 0.59 & 287 & 0.37 & 0.02 & 0.66 & 0.32 & 0.47 & 111 & 0.51 & 0.45 & 0.74 \\
    \noalign{\vskip 1.5pt}\cdashline{1-14}[1pt/1.5pt]\noalign{\vskip 1.5pt}
    $256$ keypoints & GT & 0.39 & 0.67 & 319 & 0.27 & 0.03 & 0.60 & 0.32 & 0.57 & 132 & 0.41 & 0.32 & 0.69 \\
     & recovered & 0.35 & 0.69 & 334 & 0.25 & 0.02 & 0.58 & 0.28 & 0.60 & 134 & 0.41 & 0.30 & 0.67 \\
    \noalign{\vskip 1.5pt}\cdashline{1-14}[1pt/1.5pt]\noalign{\vskip 1.5pt}
    $128$ keypoints & GT & 0.35 & 0.76 & 331 & 0.21 & 0.01 & 0.54 & 0.26 & 0.68 & 183 & 0.31 & 0.12 & 0.60 \\
     & recovered & 0.33 & 0.78 & 337 & 0.18 & 0.00 & 0.52 & 0.25 & 0.70 & 197 & 0.29 & 0.09 & 0.57 \\
    \noalign{\vskip 1.5pt}\cdashline{1-14}[1pt/1.5pt]\noalign{\vskip 1.5pt}
    $64$ keypoints & GT & 0.33 & 0.81 & 329 & 0.14 & 0.00 & 0.50 & 0.23 & 0.72 & 247 & 0.22 & 0.05 & 0.52 \\
     & recovered & 0.32 & 0.82 & 337 & 0.10 & 0.00 & 0.48 & 0.23 & 0.73 & 257 & 0.19 & 0.04 & 0.52 \\
    \midrule
    OPQ $16{\times}8$ & GT & 0.56 & 0.40 & 131 & 0.53 & 0.18 & 0.75 & 0.40 & 0.35 & 116 & 0.56 & 0.48 & 0.76 \\
     & recovered & 0.36 & 0.52 & 195 & 0.42 & 0.03 & 0.69 & 0.27 & 0.44 & 120 & 0.53 & 0.47 & 0.75 \\
    \noalign{\vskip 1.5pt}\cdashline{1-14}[1pt/1.5pt]\noalign{\vskip 1.5pt}
    OPQ $8{\times}8$ & GT & 0.54 & 0.42 & 135 & 0.49 & 0.19 & 0.73 & 0.36 & 0.39 & 118 & 0.54 & 0.49 & 0.76 \\
     & recovered & 0.32 & 0.60 & 238 & 0.34 & 0.01 & 0.63 & 0.22 & 0.55 & 135 & 0.48 & 0.35 & 0.73 \\
    \noalign{\vskip 1.5pt}\cdashline{1-14}[1pt/1.5pt]\noalign{\vskip 1.5pt}
    OPQ $4{\times}8$ & GT & 0.52 & 0.44 & 137 & 0.48 & 0.17 & 0.71 & 0.33 & 0.43 & 119 & 0.50 & 0.42 & 0.73 \\
     & recovered & 0.31 & 0.68 & 296 & 0.17 & 0.00 & 0.52 & 0.23 & 0.67 & 197 & 0.39 & 0.11 & 0.65 \\
    \bottomrule
  \end{tabular*}
  \caption{\textbf{All metrics of the benchmark of~\cite{pietrantoni2026vulnerability}}, computed with its code: SSIM,
    grayscale LPIPS, FID, and the Qwen2-VL scores OR, TR and SR (Sec.~\ref{sec:supp-metrics}), on $100$ images
    per row. COCO: ALIKED with the unadapted solver of Sec.~\ref{sec:supp-defenses}. Aachen: the LoMa query
    releases of Table~\ref{tab:supp-aachen-query}. GT and recovered: the released descriptors at the GT
    positions and at those of the solver adapted to the release; random: all descriptors at random
    positions. Every reconstruction is compared with the original image, at its inverter's output size. FID
    on $100$ images is noisy.}
  \label{tab:supp-benchmark}
\end{table*}

\begin{figure*}[t]
  \centering
  \includegraphics{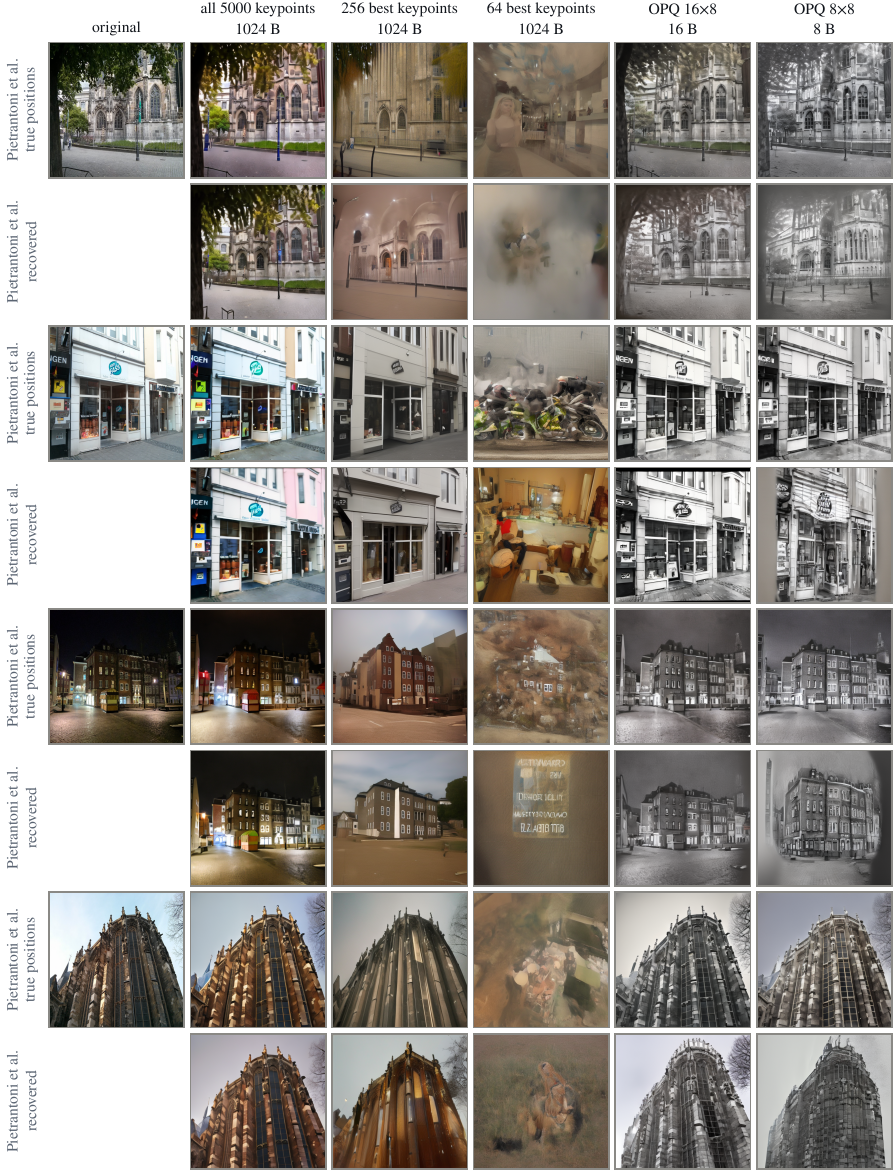}
  \caption{\textbf{Aachen queries reconstructed by the diffusion inverter.} Four queries drawn at random, under each
    release. For each query, top: GT positions with the released descriptors, the setting
    of~\cite{pietrantoni2026vulnerability}; bottom: the positions that the solver adapted to the release
    recovers. The diffusion inverter can add content that the descriptors do not carry;
    Fig.~\ref{fig:supp-aachen-queries-unet} shows the same queries with our U-Net.}
  \label{fig:supp-aachen-queries}
\end{figure*}

\begin{figure*}[t]
  \centering
  \includegraphics{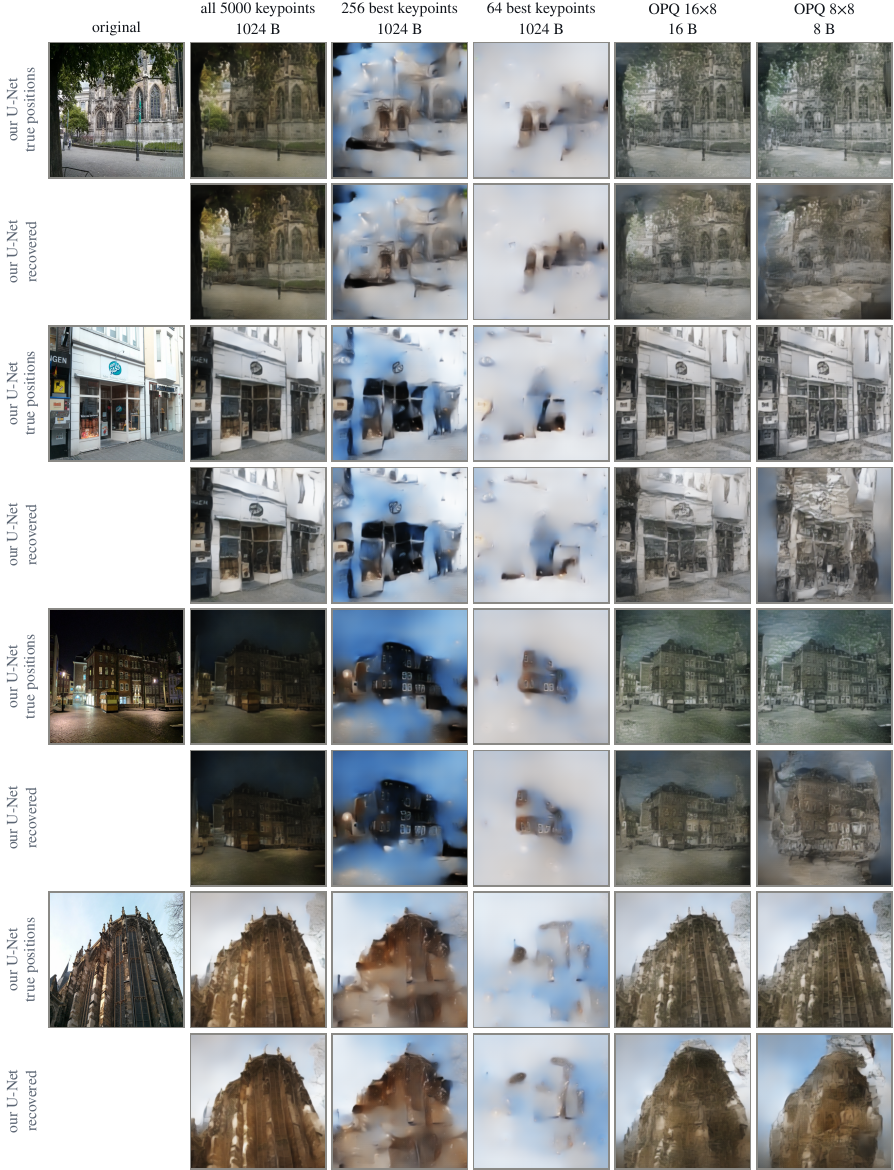}
  \caption{\textbf{The queries of Fig.~\ref{fig:supp-aachen-queries} reconstructed by our U-Net}, retrained for LoMa.
    For each query, top: GT positions; bottom: recovered positions. The U-Net renders at
    $256{\times}256$.}
  \label{fig:supp-aachen-queries-unet}
\end{figure*}

\begin{figure*}[t]
  \centering
  \includegraphics{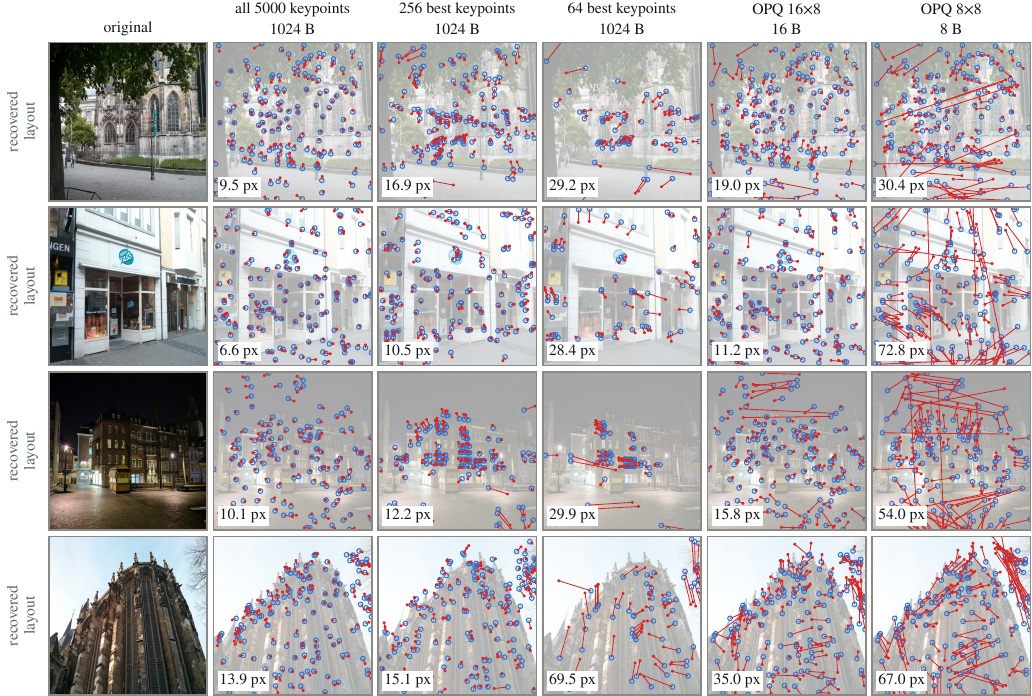}
  \caption{\textbf{Recovered layouts of the queries of Fig.~\ref{fig:supp-aachen-queries}.} GT (circles) and recovered
    (dots) positions of up to $150$ keypoints, joined by their errors; the number is the median residual of
    the release ($800$\,px frame).}
  \label{fig:supp-aachen-layouts}
\end{figure*}

Table~\ref{tab:loc} weighs what query releases cost in localization against what the attack still recovers,
on Aachen Day--Night~\cite{sattler2018benchmarking}. Here we give the setup and all results.

\customparagraph{Setup.}
We use LoMa~\cite{nordstrom2026loma} features at $800$\,px on the longer side, with up to $5{,}000$ keypoints per
query and $512$ per database image, on Aachen Day--Night v1.0. The map stores the database descriptors with $4$ bits
per value. For the localization results, the query descriptors are matched to the database by nearest neighbor search using the ratio test from~\cite{Sattler_2026_CVPR}, which does not use keypoint positions. 
Following~\cite{sattler2018benchmarking}, we report the percentage of query images localized within $25$\,cm and
$2^\circ$ of the ground truth pose. The server owns the database images, so the attacker adapts a solver on their features and
tests it on the queries (Sec.~\ref{sec:supp-defenses}). We invert $100$ queries per release, with the
diffusion inverter of~\cite{pietrantoni2026vulnerability} fine-tuned for LoMa and with our U-Net retrained for
LoMa on COCO images.

\customparagraph{Releases.}
A client can send fewer keypoints, spatially separated keypoints, or quantized descriptors. LoMa keeps its
highest-scoring keypoints, so a release of $N$ keypoints is a subset of the full one. A spaced release takes
keypoints in score order and keeps each one only if it is at least $D$\,px from those already kept. A
quantized release sends all keypoints through an OPQ codec trained on the database descriptors.
Table~\ref{tab:supp-aachen-query} reports, for each release, the unadapted query solver and the one adapted to
the same release of the database. As the release shrinks or is quantized, the unadapted solver
degrades quickly, while the adapted one is able to better recover the layout of the keypoints. %
At the same time, localization accuracy degrades as well, limiting the effectiveness of the different release strategies in terms of preventing the recovery of the keypoint position layout.

\customparagraph{What the reconstructions show.}
Figs.~\ref{fig:supp-aachen-queries} to~\ref{fig:supp-aachen-layouts} show four queries under each release.
With all keypoints, the reconstructions from recovered and from GT positions recover about the same share of
objects with both inverters. Fewer keypoints reduce what is recovered already at the GT positions, and with
the fewest neither inverter shows the original scene. The diffusion inverter then invents a plausible one,
while the U-Net draws the released descriptors in place and leaves the rest empty. The spaced releases show
the same contrast. With the two sparsest, the U-Net recovers no object, while the diffusion inverter, which
completes a street scene from little evidence, still scores slightly above its random-position level.
Quantization keeps most of the content at the GT positions, but the layout becomes harder to recover, and the
attack loses content with it. Table~\ref{tab:supp-benchmark} reports every metric of the benchmark
of~\cite{pietrantoni2026vulnerability} for these releases and for COCO.

\customparagraph{Object recovery against the GT reconstruction.}
Table~\ref{tab:loc} scores object recovery against the original image. Scored against the reconstruction from
GT positions instead, as in Table~\ref{tab:image}, the attack keeps $69\%$ of its objects with all keypoints
and $23\%$ with the strongest quantization, against $15\%$ for random positions. For sparse releases this
score is misleading, which is why Table~\ref{tab:loc} does not use it. Both reconstructions then show few real
objects, and Qwen2-VL describes them with the same generic items (``milk with bubbles'', ``spheres''). The two
sparsest spaced releases therefore score above $0.4$, although neither reconstruction contains an object of
the original image.

\section{Two paintings on a plain wall}
\label{sec:supp-copies}\suppfrozen{11}

We give the measurements behind the painting experiment of Fig.~\ref{fig:paintings} and test its
explanation in Sec.~\ref{sec:mechanism}.

\begin{table}[t]
  \centering
  \footnotesize
  \setlength{\tabcolsep}{2pt}
  \begin{tabular*}{\linewidth}{@{\extracolsep{\fill}}lccccc@{}}
    \toprule
    & & \multicolumn{2}{c}{\textbf{Different}} & \multicolumn{2}{c}{\textbf{Identical}} \\
    \cmidrule(lr){3-4}\cmidrule(l){5-6}
    \textbf{Gap (px)} & \textbf{GT} & \textbf{Side} & \textbf{Stacked} & \textbf{Side} & \textbf{Stacked} \\
    \midrule
    $0$ & $62$ & $59$ & $67$ & $58$ & $66$ \\
    $20$ & $83$ & $83$ & $86$ & $84$ & $87$ \\
    $40$ & $104$ & $104$ & $111$ & $36$ & $86$ \\
    $60$ & $123$ & $119$ & $128$ & $1$ & $28$ \\
    $80$ & $144$ & $72$ & $64$ & $1$ & $4$ \\
    $100$ & $164$ & $37$ & $15$ & $1$ & $1$ \\
    $120$ & $184$ & $23$ & $8$ & $1$ & $1$ \\
    $160$ & $224$ & $8$ & $5$ & $1$ & $1$ \\
    \bottomrule
  \end{tabular*}
  \caption{\textbf{Two paintings far from the image border} (Fig.~\ref{fig:paintings}c). Median GT and recovered
    distance between the paintings' keypoint centroids (px) against the gap between their edges, on a
    $1024{\times}1024$ wall, $100$ pairs per condition.}
  \label{tab:supp-paintings-far}
\end{table}

\customparagraph{Far from the border.}
Table~\ref{tab:supp-paintings-far} lists the numbers behind Fig.~\ref{fig:paintings}c. The paintings hang at
the center of a $1024{\times}1024$ wall, far from every edge, and the features are extracted at twice the
wall's size, as for the $512$\,px wall, so the paintings keep their size in pixels. At small gaps every pair
keeps its distance. Different paintings keep it up to a gap of about $50$\,px and nearly coincide from about
$100$\,px, while identical copies merge earlier. A larger wall gives the same result. On the $512$\,px wall
the image border helps. There, side-by-side copies stay apart up to $160$\,px, and stacked copies only near
the top edge.

\customparagraph{Why stacked copies merge.}
Two identical copies differ only in where they lie in the image, and ALIKED's descriptors change with that
position, a little more horizontally than vertically. At the same gap, stacked copies are therefore more alike
than side-by-side ones, as the paper reports. To see how much difference the solver needs, we keep the keypoints
that have an exact counterpart on the other copy and scale the difference between the two copies' descriptors.
Side-by-side copies differ by $0.018$ and stay apart at that difference. Stacked copies differ by $0.013$ and need
$0.030$, almost twice as much. Stacked copies thus both differ less and need a larger difference to stay apart.
Most COCO images are wider than tall, which may explain the second effect, although we have not tested it.

\begin{table}[t]
  \centering
  \footnotesize
  \setlength{\tabcolsep}{4pt}
  \begin{tabular*}{\linewidth}{@{\extracolsep{\fill}}lcccc@{}}
    \toprule
    & \multicolumn{2}{c}{\textbf{Side by side}} & \multicolumn{2}{c}{\textbf{Stacked}} \\
    \cmidrule(lr){2-3}\cmidrule(l){4-5}
    \textbf{Gap (px)} & \textbf{Both} & \textbf{Alone} & \textbf{Both} & \textbf{Alone} \\
    \midrule
    $0$ & $2$ & $53$ & $5$ & $57$ \\
    $32$ & $3$ & $55$ & $7$ & $88$ \\
    $64$ & $14$ & $31$ & $13$ & $110$ \\
    $96$ & $12$ & $20$ & $15$ & $92$ \\
    $128$ & $16$ & $20$ & $12$ & $44$ \\
    $160$ & $21$ & $29$ & $14$ & $24$ \\
    $192$ & $43$ & $44$ & $32$ & $34$ \\
    \bottomrule
  \end{tabular*}
  \caption{\textbf{Each painting described without the other.} Median error of the recovered offset between two
    different paintings on the $512$\,px wall ($100$ pairs), with descriptors computed in the scene with
    both paintings (both) or in an image without the other painting, at the same keypoints (alone).}
  \label{tab:supp-paintings-alone}
\end{table}

\customparagraph{Each painting described without the other.}
If nearby paintings are placed through what their descriptors see of each other, taking that view away should
break the placement. We keep the keypoints of the scene with both paintings on the $512$\,px wall, but compute
each painting's descriptors in an image without the other one (Table~\ref{tab:supp-paintings-alone}). At small
gaps the error jumps, and touching paintings are recovered almost on top of each other. At large gaps the
difference is small, and the image border keeps the paintings apart.

Plain walls are outside the training distribution, and the $1024$\,px wall is larger than any training image.
Even there, close pairs are recovered at their true distance.

\section{Documents}
\label{sec:supp-documents}\suppfrozen{12}

Sec.~\ref{sec:exp-inversion} reads text from reconstructed FUNSD forms~\cite{jaume2019funsd}. The ALIKED
solver recovers their layout without retraining (Table~\ref{tab:main}), but Qwen2-VL reads almost no words in the
reconstructions of our COCO U-Net and of the diffusion inverter, even from GT positions and even after fine-tuning
the diffusion inverter on documents. We therefore train a document U-Net with the recipe of Sec.~\ref{sec:supp-impl} on RVL-CDIP
pages~\cite{harley2015rvlcdip}, after removing the FUNSD forms and their near-duplicates. It uses a finer grid
than the COCO U-Net, since pages resampled to the coarser grid already lose about a fifth of their readable
words.

\customparagraph{Measuring text.}
Qwen2-VL~\cite{wang2024qwen2vl} transcribes each image, and word recall is the share of the form's distinct
words of at least three letters or digits that appear in the transcription. Scoring a form's words against
the transcription of another form gives the chance level, which is above zero because forms share words such
as ``date''.

\begin{table}[t]
  \centering
  \footnotesize
  \setlength{\tabcolsep}{2pt}
  \begin{tabular*}{\linewidth}{@{\extracolsep{\fill}}lccc@{}}
    \toprule
    \textbf{Positions} & \textbf{Error (px)} $\downarrow$ & \textbf{Words} $\uparrow$ & \textbf{Chance} \\
    \midrule
    Original page & -- & $0.78$ & $0.10$ \\
    GT & $0$ & $0.50$ & $0.08$ \\
    Recovered & $5.9$ & $0.15$ & $0.05$ \\
    Random & $207.1$ & $0.02$ & $0.02$ \\
    \bottomrule
  \end{tabular*}
  \caption{\textbf{Words read by the document U-Net} on the $50$ FUNSD test forms. Error: median layout residual;
    words: median word recall; chance: recall against another form's transcription.}
  \label{tab:supp-documents}
\end{table}

\customparagraph{Result.}
In the document U-Net's reconstructions, Qwen2-VL reads half of the words at GT positions and $15\%$ at recovered
ones (Table~\ref{tab:supp-documents}). The page structure, the rules and large print survive
(Fig.~\ref{fig:documents}), but small print needs a more precise layout than the attack provides
(Sec.~\ref{sec:supp-precision}).

\section{Layout precision}
\label{sec:supp-precision}\suppfrozen{13}

To find how much position error the content tolerates, we move the GT positions on purpose and invert the
moved layout. Either each keypoint moves independently by a Gaussian offset, or nearby keypoints move together
along a smooth random displacement field. The second setting is closer to the attack, whose errors are
strongly correlated between nearby keypoints. The error of a moved layout is measured like the attack's.
Objects are scored on $100$ COCO test images with our U-Net, words on the $50$ FUNSD forms with the document
U-Net.

\begin{table}[t]
  \centering
  \footnotesize
  \setlength{\tabcolsep}{2pt}
  \begin{tabular*}{\linewidth}{@{\extracolsep{\fill}}lccccc@{}}
    \toprule
    \multicolumn{3}{c}{\textbf{Moved independently}} & \multicolumn{3}{c}{\textbf{Moved together}} \\
    \cmidrule(lr){1-3}\cmidrule(lr){4-6}
    \textbf{Error (px)} & \textbf{OR} $\uparrow$ & \textbf{Words} $\uparrow$ & \textbf{Error (px)} & \textbf{OR} $\uparrow$ & \textbf{Words} $\uparrow$ \\
    \midrule
    $1.2$ & $0.52$ & $0.09$ & $2.4$ & $0.54$ & $0.49$ \\
    $2.3$ & $0.51$ & $0.00$ & $4.8$ & $0.52$ & $0.41$ \\
    $4.7$ & $0.38$ & $0.00$ & $9.6$ & $0.48$ & $0.29$ \\
    $9.4$ & $0.26$ & $0.01$ & $19.2$ & $0.40$ & $0.10$ \\
    $18.8$ & $0.16$ & $0.03$ & $38.4$ & $0.23$ & $0.01$ \\
    $37.6$ & $0.07$ & $0.03$ & $76.9$ & $0.13$ & $0.02$ \\
    \bottomrule
  \end{tabular*}
  \caption{\textbf{Content against position error.} Error: median distance of the moved keypoints from their GT positions
    (px, $512$\,px frame; the FUNSD errors differ from these by less than $1$\,px). Moved together: white noise smoothed with a Gaussian of $46$\,px. OR on $100$ COCO test images; words: median word
    recall on the $50$ FUNSD forms. GT positions give OR $0.56$ and words $0.50$, random positions $0.05$
    and $0.02$. The attack on FUNSD reaches $5.9$\,px and words $0.15$.}
  \label{tab:supp-precision}
\end{table}

\customparagraph{Result.}
Objects tolerate a few pixels of either kind of error (Table~\ref{tab:supp-precision}). Text does not tolerate
independent errors. Moving each keypoint by about $2$\,px already leaves no word readable, while moving
neighbors together by twice as much keeps many of the words. Some words survive the attack only because its
neighboring keypoints move together.